%% file: icml26_lgt.tex
\documentclass{article}

\usepackage{microtype}
\usepackage{graphicx}
\usepackage{subcaption}
\usepackage{booktabs} 

\usepackage{hyperref}

\usepackage[preprint]{icml2026}

\usepackage{amsmath}
\usepackage{amssymb}
\usepackage{mathtools}
\usepackage{amsthm}
\usepackage[utf8]{inputenc}
\usepackage[T1]{fontenc}
\usepackage[dvipsnames,svgnames,x11names]{xcolor}
\usepackage{adjustbox}
\usepackage{fontawesome5}
\usepackage[most,skins,breakable,raster]{tcolorbox}
\usepackage{booktabs}
\usepackage{multirow}
\usepackage{wrapfig} 
\usepackage{tabularx}
\usepackage{pifont,xcolor}
\usepackage{xspace}
\usepackage{subcaption}

\usepackage[capitalize,noabbrev]{cleveref}

\theoremstyle{plain}

\theoremstyle{definition}

\theoremstyle{remark}

\usepackage[textsize=tiny]{todonotes}

\icmltitlerunning{Long Grounded Thoughts: Synthesizing  Visual Problems and Reasoning Chains at Scale}

\newtcolorbox{promptbox}[1][]{
    enhanced,
    colback=blue!5!white,
    colframe=blue!75!black,
    fonttitle=\bfseries,
    title=Prompt Task,
    attach boxed title to top left={yshift=-2mm, xshift=3mm},
    boxed title style={colback=blue!75!black},
    #1
}

\newtcolorbox{principlesbox}[2][]{
    enhanced,
    breakable,
    colback=green!5!white,
    colframe=green!60!black,
    fonttitle=\bfseries,
    title=#2,
    attach boxed title to top left={yshift=-2mm, xshift=3mm},
    boxed title style={colback=green!60!black},
    #1
}

\newtcolorbox{instructionbox}[2][]{
    enhanced,
    breakable,
    colback=orange!5!white,
    colframe=orange!80!black,
    fonttitle=\bfseries,
    title=#2,
    attach boxed title to top left={yshift=-2mm, xshift=3mm},
    boxed title style={colback=orange!80!black},
    #1
}

\newtcolorbox{criticalbox}[2][]{
    enhanced,
    breakable,
    colback=red!5!white,
    colframe=red!75!black,
    fonttitle=\bfseries,
    title=#2,
    attach boxed title to top left={yshift=-2mm, xshift=3mm},
    boxed title style={colback=red!75!black},
    #1
}

\newtcolorbox{databox}[2][]{
    enhanced,
    breakable,
    colback=black!5!white,
    colframe=black!60!black,
    fonttitle=\bfseries\ttfamily,
    title=#2,
    #1
}

\newtcolorbox{qualitativecomparison}[1][]{
    enhanced,
    breakable,
    colback=black!2!white,
    colframe=black!60!black,
    fonttitle=\bfseries,
    title=Qualitative Comparison: SFT-DPO vs. SFT Baseline,
    attach boxed title to top left={yshift=-2mm, xshift=3mm},
    boxed title style={colback=black!60!black},
    #1
}

\newtcolorbox{dataqualitybox}[1][]{
    enhanced,
    breakable,
    colback=Goldenrod!5!white,
    colframe=Goldenrod!85!black,
    fonttitle=\bfseries,
    attach boxed title to top left={yshift=-2mm, xshift=3mm},
    boxed title style={colback=Goldenrod!85!black},
    #1
}
\definecolor{ngreen}{HTML}{76B900}

\newif\ifrebuttalmode

\rebuttalmodefalse

\ifrebuttalmode
    \usepackage{xcolor} 
    \newcommand{\rebuttal}[1]{\textcolor{blue}{#1}}
    \newcommand{\new}[1]{\textcolor{blue}{#1}}
\else
    \newcommand{\rebuttal}[1]{#1}
    \newcommand{\new}[1]{#1}
\fi

\begin{document}

\twocolumn[
  \icmltitle{
  Long Grounded Thoughts: Synthesizing  Visual Problems\\and  Reasoning Chains at Scale
  }



  \icmlsetsymbol{equal}{*}

  \begin{icmlauthorlist}
  \icmlauthor{David Acuna}{nvidia,equal}
  \icmlauthor{Chao-Han Huck Yang}{nvidia,equal}
  \icmlauthor{Yuntian Deng}{nvidia,uwaterloo,equal}
  \icmlauthor{Jaehun Jung}{nvidia,equal}
  \icmlauthor{Ximing Lu}{nvidia,equal}
  \icmlauthor{Prithviraj Ammanabrolu}{nvidia,ucsd}
  \icmlauthor{Hyunwoo Kim}{nvidia}
  \icmlauthor{Yuan-Hong Liao}{uoft}
  \icmlauthor{Yejin Choi}{nvidia}
\end{icmlauthorlist}

  \icmlaffiliation{nvidia}{NVIDIA}
\icmlaffiliation{uoft}{University of Toronto}
\icmlaffiliation{uwaterloo}{University of Waterloo}
\icmlaffiliation{ucsd}{UCSD}
\icmlcorrespondingauthor{David Acuna}{dacunamarrer@nvidia.com}

  \icmlkeywords{Machine Learning, ICML}

  \vskip 0.3in
]
\newcommand{\fix}{\marginpar{FIX}}
\newcommand{\ourdataset}{Long Grounded Thoughts\xspace}
\newcommand\jaehun[1]{{\color{teal}[#1]$_{Jaehun}$}}

\input{math_commands}
\input{macro}


\printAffiliationsAndNotice{}  

\begin{abstract}

Despite rapid progress, multimodal reasoning still lacks a systematic approach to synthesize large-scale vision-centric datasets beyond visual math.
We introduce a framework able to synthesize \textit{vision-centric} problems spanning diverse levels of complexity, and the resulting dataset with over 1M high-quality problems including: reasoning traces, preference data, and instruction prompts supporting SFT, offline and online RL. 
Our vision-centric  synthesis framework uses a two-stage process focusing on: (1) generating diverse verifiable questions from existing images at scale, and (2) creating complex compositional visual problems by merging simpler questions. 
%
Remarkably, 
finetuning Qwen2.5-VL-7B on our data outperforms existing open-data baselines across evaluated vision-centric benchmarks, and our best configurations match or surpass strong closed-data models 
such as MiMo-VL-7B-RL 
on V*Bench, CV-Bench and MMStar-V.
Notably, despite being entirely vision-centric, our data transfers positively to text-only reasoning (MMLU-Pro, +3.7\%) and audio reasoning (MMAU, +1.32\%), demonstrating its  effectiveness. \rebuttal{Similarly, despite  containing no  embodied visual data, we observe notable gains (NiEH, +8.8\%)  when evaluating open-ended 
embodied QA}. \new{Lastly, we use our data to comprehensively analyze at scale (1M+) the entire VLM post-training pipeline
showing that (i) SFT on high-quality data with cognitive behaviors on reasoning traces is essential to scale online RL, (ii) offline RL could match online RL’s performance while disaggregating compute demands, and, (iii)  SFT on high quality data also improve out-of-domain, cross-modality transfer.}

\end{abstract}

\input{sections/1-intro}
\input{sections/2-method}
\input{sections/3-exp}
\input{sections/related}

\input{sections/4-discussion}

\newpage
\input{sections/requirements}


\bibliography{icml2026_conference}
\bibliographystyle{icml2026}


\newpage
\appendix
\onecolumn
\section*{Appendix}

\input{sections/appendix_simplified}


\end{document}

%% file: math_commands.tex


\newcommand{\figleft}{{\em (Left)}}
\newcommand{\figcenter}{{\em (Center)}}
\newcommand{\figright}{{\em (Right)}}
\newcommand{\figtop}{{\em (Top)}}
\newcommand{\figbottom}{{\em (Bottom)}}
\newcommand{\captiona}{{\em (a)}}
\newcommand{\captionb}{{\em (b)}}
\newcommand{\captionc}{{\em (c)}}
\newcommand{\captiond}{{\em (d)}}

\newcommand{\newterm}[1]{{\bf #1}}

\def\figref#1{figure~\ref{#1}}
\def\Figref#1{Figure~\ref{#1}}
\def\twofigref#1#2{figures \ref{#1} and \ref{#2}}
\def\quadfigref#1#2#3#4{figures \ref{#1}, \ref{#2}, \ref{#3} and \ref{#4}}
\def\secref#1{section~\ref{#1}}
\def\Secref#1{Section~\ref{#1}}
\def\twosecrefs#1#2{sections \ref{#1} and \ref{#2}}
\def\secrefs#1#2#3{sections \ref{#1}, \ref{#2} and \ref{#3}}
\def\eqref#1{equation~\ref{#1}}
\def\Eqref#1{Equation~\ref{#1}}
\def\plaineqref#1{\ref{#1}}
\def\chapref#1{chapter~\ref{#1}}
\def\Chapref#1{Chapter~\ref{#1}}
\def\rangechapref#1#2{chapters\ref{#1}--\ref{#2}}
\def\algref#1{algorithm~\ref{#1}}
\def\Algref#1{Algorithm~\ref{#1}}
\def\twoalgref#1#2{algorithms \ref{#1} and \ref{#2}}
\def\Twoalgref#1#2{Algorithms \ref{#1} and \ref{#2}}
\def\partref#1{part~\ref{#1}}
\def\Partref#1{Part~\ref{#1}}
\def\twopartref#1#2{parts \ref{#1} and \ref{#2}}

\def\ceil#1{\lceil #1 \rceil}
\def\floor#1{\lfloor #1 \rfloor}
\def\1{\bm{1}}
\newcommand{\train}{\mathcal{D}}
\newcommand{\valid}{\mathcal{D_{\mathrm{valid}}}}
\newcommand{\test}{\mathcal{D_{\mathrm{test}}}}

\def\eps{{\epsilon}}

\def\reta{{\textnormal{$\eta$}}}
\def\ra{{\textnormal{a}}}
\def\rb{{\textnormal{b}}}
\def\rc{{\textnormal{c}}}
\def\rd{{\textnormal{d}}}
\def\re{{\textnormal{e}}}
\def\rf{{\textnormal{f}}}
\def\rg{{\textnormal{g}}}
\def\rh{{\textnormal{h}}}
\def\ri{{\textnormal{i}}}
\def\rj{{\textnormal{j}}}
\def\rk{{\textnormal{k}}}
\def\rl{{\textnormal{l}}}
\def\rn{{\textnormal{n}}}
\def\ro{{\textnormal{o}}}
\def\rp{{\textnormal{p}}}
\def\rq{{\textnormal{q}}}
\def\rr{{\textnormal{r}}}
\def\rs{{\textnormal{s}}}
\def\rt{{\textnormal{t}}}
\def\ru{{\textnormal{u}}}
\def\rv{{\textnormal{v}}}
\def\rw{{\textnormal{w}}}
\def\rx{{\textnormal{x}}}
\def\ry{{\textnormal{y}}}
\def\rz{{\textnormal{z}}}

\def\rvepsilon{{\mathbf{\epsilon}}}
\def\rvtheta{{\mathbf{\theta}}}
\def\rva{{\mathbf{a}}}
\def\rvb{{\mathbf{b}}}
\def\rvc{{\mathbf{c}}}
\def\rvd{{\mathbf{d}}}
\def\rve{{\mathbf{e}}}
\def\rvf{{\mathbf{f}}}
\def\rvg{{\mathbf{g}}}
\def\rvh{{\mathbf{h}}}
\def\rvu{{\mathbf{i}}}
\def\rvj{{\mathbf{j}}}
\def\rvk{{\mathbf{k}}}
\def\rvl{{\mathbf{l}}}
\def\rvm{{\mathbf{m}}}
\def\rvn{{\mathbf{n}}}
\def\rvo{{\mathbf{o}}}
\def\rvp{{\mathbf{p}}}
\def\rvq{{\mathbf{q}}}
\def\rvr{{\mathbf{r}}}
\def\rvs{{\mathbf{s}}}
\def\rvt{{\mathbf{t}}}
\def\rvu{{\mathbf{u}}}
\def\rvv{{\mathbf{v}}}
\def\rvw{{\mathbf{w}}}
\def\rvx{{\mathbf{x}}}
\def\rvy{{\mathbf{y}}}
\def\rvz{{\mathbf{z}}}

\def\erva{{\textnormal{a}}}
\def\ervb{{\textnormal{b}}}
\def\ervc{{\textnormal{c}}}
\def\ervd{{\textnormal{d}}}
\def\erve{{\textnormal{e}}}
\def\ervf{{\textnormal{f}}}
\def\ervg{{\textnormal{g}}}
\def\ervh{{\textnormal{h}}}
\def\ervi{{\textnormal{i}}}
\def\ervj{{\textnormal{j}}}
\def\ervk{{\textnormal{k}}}
\def\ervl{{\textnormal{l}}}
\def\ervm{{\textnormal{m}}}
\def\ervn{{\textnormal{n}}}
\def\ervo{{\textnormal{o}}}
\def\ervp{{\textnormal{p}}}
\def\ervq{{\textnormal{q}}}
\def\ervr{{\textnormal{r}}}
\def\ervs{{\textnormal{s}}}
\def\ervt{{\textnormal{t}}}
\def\ervu{{\textnormal{u}}}
\def\ervv{{\textnormal{v}}}
\def\ervw{{\textnormal{w}}}
\def\ervx{{\textnormal{x}}}
\def\ervy{{\textnormal{y}}}
\def\ervz{{\textnormal{z}}}

\def\rmA{{\mathbf{A}}}
\def\rmB{{\mathbf{B}}}
\def\rmC{{\mathbf{C}}}
\def\rmD{{\mathbf{D}}}
\def\rmE{{\mathbf{E}}}
\def\rmF{{\mathbf{F}}}
\def\rmG{{\mathbf{G}}}
\def\rmH{{\mathbf{H}}}
\def\rmI{{\mathbf{I}}}
\def\rmJ{{\mathbf{J}}}
\def\rmK{{\mathbf{K}}}
\def\rmL{{\mathbf{L}}}
\def\rmM{{\mathbf{M}}}
\def\rmN{{\mathbf{N}}}
\def\rmO{{\mathbf{O}}}
\def\rmP{{\mathbf{P}}}
\def\rmQ{{\mathbf{Q}}}
\def\rmR{{\mathbf{R}}}
\def\rmS{{\mathbf{S}}}
\def\rmT{{\mathbf{T}}}
\def\rmU{{\mathbf{U}}}
\def\rmV{{\mathbf{V}}}
\def\rmW{{\mathbf{W}}}
\def\rmX{{\mathbf{X}}}
\def\rmY{{\mathbf{Y}}}
\def\rmZ{{\mathbf{Z}}}

\def\ermA{{\textnormal{A}}}
\def\ermB{{\textnormal{B}}}
\def\ermC{{\textnormal{C}}}
\def\ermD{{\textnormal{D}}}
\def\ermE{{\textnormal{E}}}
\def\ermF{{\textnormal{F}}}
\def\ermG{{\textnormal{G}}}
\def\ermH{{\textnormal{H}}}
\def\ermI{{\textnormal{I}}}
\def\ermJ{{\textnormal{J}}}
\def\ermK{{\textnormal{K}}}
\def\ermL{{\textnormal{L}}}
\def\ermM{{\textnormal{M}}}
\def\ermN{{\textnormal{N}}}
\def\ermO{{\textnormal{O}}}
\def\ermP{{\textnormal{P}}}
\def\ermQ{{\textnormal{Q}}}
\def\ermR{{\textnormal{R}}}
\def\ermS{{\textnormal{S}}}
\def\ermT{{\textnormal{T}}}
\def\ermU{{\textnormal{U}}}
\def\ermV{{\textnormal{V}}}
\def\ermW{{\textnormal{W}}}
\def\ermX{{\textnormal{X}}}
\def\ermY{{\textnormal{Y}}}
\def\ermZ{{\textnormal{Z}}}

\def\vzero{{\bm{0}}}
\def\vone{{\bm{1}}}
\def\vmu{{\bm{\mu}}}
\def\vtheta{{\bm{\theta}}}
\def\va{{\bm{a}}}
\def\vb{{\bm{b}}}
\def\vc{{\bm{c}}}
\def\vd{{\bm{d}}}
\def\ve{{\bm{e}}}
\def\vf{{\bm{f}}}
\def\vg{{\bm{g}}}
\def\vh{{\bm{h}}}
\def\vi{{\bm{i}}}
\def\vj{{\bm{j}}}
\def\vk{{\bm{k}}}
\def\vl{{\bm{l}}}
\def\vm{{\bm{m}}}
\def\vn{{\bm{n}}}
\def\vo{{\bm{o}}}
\def\vp{{\bm{p}}}
\def\vq{{\bm{q}}}
\def\vr{{\bm{r}}}
\def\vs{{\bm{s}}}
\def\vt{{\bm{t}}}
\def\vu{{\bm{u}}}
\def\vv{{\bm{v}}}
\def\vw{{\bm{w}}}
\def\vx{{\bm{x}}}
\def\vy{{\bm{y}}}
\def\vz{{\bm{z}}}

\def\evalpha{{\alpha}}
\def\evbeta{{\beta}}
\def\evepsilon{{\epsilon}}
\def\evlambda{{\lambda}}
\def\evomega{{\omega}}
\def\evmu{{\mu}}
\def\evpsi{{\psi}}
\def\evsigma{{\sigma}}
\def\evtheta{{\theta}}
\def\eva{{a}}
\def\evb{{b}}
\def\evc{{c}}
\def\evd{{d}}
\def\eve{{e}}
\def\evf{{f}}
\def\evg{{g}}
\def\evh{{h}}
\def\evi{{i}}
\def\evj{{j}}
\def\evk{{k}}
\def\evl{{l}}
\def\evm{{m}}
\def\evn{{n}}
\def\evo{{o}}
\def\evp{{p}}
\def\evq{{q}}
\def\evr{{r}}
\def\evs{{s}}
\def\evt{{t}}
\def\evu{{u}}
\def\evv{{v}}
\def\evw{{w}}
\def\evx{{x}}
\def\evy{{y}}
\def\evz{{z}}

\def\mA{{\bm{A}}}
\def\mB{{\bm{B}}}
\def\mC{{\bm{C}}}
\def\mD{{\bm{D}}}
\def\mE{{\bm{E}}}
\def\mF{{\bm{F}}}
\def\mG{{\bm{G}}}
\def\mH{{\bm{H}}}
\def\mI{{\bm{I}}}
\def\mJ{{\bm{J}}}
\def\mK{{\bm{K}}}
\def\mL{{\bm{L}}}
\def\mM{{\bm{M}}}
\def\mN{{\bm{N}}}
\def\mO{{\bm{O}}}
\def\mP{{\bm{P}}}
\def\mQ{{\bm{Q}}}
\def\mR{{\bm{R}}}
\def\mS{{\bm{S}}}
\def\mT{{\bm{T}}}
\def\mU{{\bm{U}}}
\def\mV{{\bm{V}}}
\def\mW{{\bm{W}}}
\def\mX{{\bm{X}}}
\def\mY{{\bm{Y}}}
\def\mZ{{\bm{Z}}}
\def\mBeta{{\bm{\beta}}}
\def\mPhi{{\bm{\Phi}}}
\def\mLambda{{\bm{\Lambda}}}
\def\mSigma{{\bm{\Sigma}}}

\newcommand{\tens}[1]{\bm{\mathsfit{#1}}}
\def\tA{{\tens{A}}}
\def\tB{{\tens{B}}}
\def\tC{{\tens{C}}}
\def\tD{{\tens{D}}}
\def\tE{{\tens{E}}}
\def\tF{{\tens{F}}}
\def\tG{{\tens{G}}}
\def\tH{{\tens{H}}}
\def\tI{{\tens{I}}}
\def\tJ{{\tens{J}}}
\def\tK{{\tens{K}}}
\def\tL{{\tens{L}}}
\def\tM{{\tens{M}}}
\def\tN{{\tens{N}}}
\def\tO{{\tens{O}}}
\def\tP{{\tens{P}}}
\def\tQ{{\tens{Q}}}
\def\tR{{\tens{R}}}
\def\tS{{\tens{S}}}
\def\tT{{\tens{T}}}
\def\tU{{\tens{U}}}
\def\tV{{\tens{V}}}
\def\tW{{\tens{W}}}
\def\tX{{\tens{X}}}
\def\tY{{\tens{Y}}}
\def\tZ{{\tens{Z}}}

\def\gA{{\mathcal{A}}}
\def\gB{{\mathcal{B}}}
\def\gC{{\mathcal{C}}}
\def\gD{{\mathcal{D}}}
\def\gE{{\mathcal{E}}}
\def\gF{{\mathcal{F}}}
\def\gG{{\mathcal{G}}}
\def\gH{{\mathcal{H}}}
\def\gI{{\mathcal{I}}}
\def\gJ{{\mathcal{J}}}
\def\gK{{\mathcal{K}}}
\def\gL{{\mathcal{L}}}
\def\gM{{\mathcal{M}}}
\def\gN{{\mathcal{N}}}
\def\gO{{\mathcal{O}}}
\def\gP{{\mathcal{P}}}
\def\gQ{{\mathcal{Q}}}
\def\gR{{\mathcal{R}}}
\def\gS{{\mathcal{S}}}
\def\gT{{\mathcal{T}}}
\def\gU{{\mathcal{U}}}
\def\gV{{\mathcal{V}}}
\def\gW{{\mathcal{W}}}
\def\gX{{\mathcal{X}}}
\def\gY{{\mathcal{Y}}}
\def\gZ{{\mathcal{Z}}}

\def\sA{{\mathbb{A}}}
\def\sB{{\mathbb{B}}}
\def\sC{{\mathbb{C}}}
\def\sD{{\mathbb{D}}}
\def\sF{{\mathbb{F}}}
\def\sG{{\mathbb{G}}}
\def\sH{{\mathbb{H}}}
\def\sI{{\mathbb{I}}}
\def\sJ{{\mathbb{J}}}
\def\sK{{\mathbb{K}}}
\def\sL{{\mathbb{L}}}
\def\sM{{\mathbb{M}}}
\def\sN{{\mathbb{N}}}
\def\sO{{\mathbb{O}}}
\def\sP{{\mathbb{P}}}
\def\sQ{{\mathbb{Q}}}
\def\sR{{\mathbb{R}}}
\def\sS{{\mathbb{S}}}
\def\sT{{\mathbb{T}}}
\def\sU{{\mathbb{U}}}
\def\sV{{\mathbb{V}}}
\def\sW{{\mathbb{W}}}
\def\sX{{\mathbb{X}}}
\def\sY{{\mathbb{Y}}}
\def\sZ{{\mathbb{Z}}}

\def\emLambda{{\Lambda}}
\def\emA{{A}}
\def\emB{{B}}
\def\emC{{C}}
\def\emD{{D}}
\def\emE{{E}}
\def\emF{{F}}
\def\emG{{G}}
\def\emH{{H}}
\def\emI{{I}}
\def\emJ{{J}}
\def\emK{{K}}
\def\emL{{L}}
\def\emM{{M}}
\def\emN{{N}}
\def\emO{{O}}
\def\emP{{P}}
\def\emQ{{Q}}
\def\emR{{R}}
\def\emS{{S}}
\def\emT{{T}}
\def\emU{{U}}
\def\emV{{V}}
\def\emW{{W}}
\def\emX{{X}}
\def\emY{{Y}}
\def\emZ{{Z}}
\def\emSigma{{\Sigma}}

\newcommand{\etens}[1]{\mathsfit{#1}}
\def\etLambda{{\etens{\Lambda}}}
\def\etA{{\etens{A}}}
\def\etB{{\etens{B}}}
\def\etC{{\etens{C}}}
\def\etD{{\etens{D}}}
\def\etE{{\etens{E}}}
\def\etF{{\etens{F}}}
\def\etG{{\etens{G}}}
\def\etH{{\etens{H}}}
\def\etI{{\etens{I}}}
\def\etJ{{\etens{J}}}
\def\etK{{\etens{K}}}
\def\etL{{\etens{L}}}
\def\etM{{\etens{M}}}
\def\etN{{\etens{N}}}
\def\etO{{\etens{O}}}
\def\etP{{\etens{P}}}
\def\etQ{{\etens{Q}}}
\def\etR{{\etens{R}}}
\def\etS{{\etens{S}}}
\def\etT{{\etens{T}}}
\def\etU{{\etens{U}}}
\def\etV{{\etens{V}}}
\def\etW{{\etens{W}}}
\def\etX{{\etens{X}}}
\def\etY{{\etens{Y}}}
\def\etZ{{\etens{Z}}}

\newcommand{\pdata}{p_{\rm{data}}}
\newcommand{\ptrain}{\hat{p}_{\rm{data}}}
\newcommand{\Ptrain}{\hat{P}_{\rm{data}}}
\newcommand{\pmodel}{p_{\rm{model}}}
\newcommand{\Pmodel}{P_{\rm{model}}}
\newcommand{\ptildemodel}{\tilde{p}_{\rm{model}}}
\newcommand{\pencode}{p_{\rm{encoder}}}
\newcommand{\pdecode}{p_{\rm{decoder}}}
\newcommand{\precons}{p_{\rm{reconstruct}}}

\newcommand{\laplace}{\mathrm{Laplace}} 

\newcommand{\E}{\mathbb{E}}
\newcommand{\Ls}{\mathcal{L}}
\newcommand{\R}{\mathbb{R}}
\newcommand{\emp}{\tilde{p}}
\newcommand{\lr}{\alpha}
\newcommand{\reg}{\lambda}
\newcommand{\rect}{\mathrm{rectifier}}
\newcommand{\softmax}{\mathrm{softmax}}
\newcommand{\sigmoid}{\sigma}
\newcommand{\softplus}{\zeta}
\newcommand{\KL}{D_{\mathrm{KL}}}
\newcommand{\Var}{\mathrm{Var}}
\newcommand{\standarderror}{\mathrm{SE}}
\newcommand{\Cov}{\mathrm{Cov}}
\newcommand{\normlzero}{L^0}
\newcommand{\normlone}{L^1}
\newcommand{\normltwo}{L^2}
\newcommand{\normlp}{L^p}
\newcommand{\normmax}{L^\infty}

\newcommand{\parents}{Pa} 

\let\ab\allowbreak

%% file: macro.tex
\newcommand{\onedot}{.\xspace}

\def\eg{\emph{e.g}\onedot} 
\def\Eg{\emph{E.g}\onedot}
\def\ie{\emph{i.e}\onedot} 
\def\Ie{\emph{I.e}\onedot}
\def\cf{\emph{cf}\onedot} 
\def\Cf{\emph{Cf}\onedot}
\def\etc{\emph{etc}\onedot} 
\def\vs{\emph{vs}\onedot}
\def\wrt{w.r.t\onedot} 
\def\dof{d.o.f\onedot}
\def\iid{i.i.d\onedot} 
\def\wolog{w.l.o.g\onedot}
\def\etal{\emph{et al}\onedot}


\newcommand{\longcot}{LongPerceptualThoughts\xspace}
\newcommand{\NA}{\textemdash} 
\newcommand{\cmark}{\textcolor{green!55!black}{\ding{51}}} 
\newcommand{\xmark}{\textcolor{red!70!black}{\ding{55}}}    
\newcommand{\model}{\mathcal{M}}           
\newcommand{\expansionprocess}{\mathcal{A}}           
\newcommand{\VLM}{\mathcal{M}_\text{VLM}}
\newcommand{\LLM}{\mathcal{M}_\text{LLM}}
\newcommand{\RLLM}{\mathcal{M}_\text{Reason}}
\newcommand{\objectmetadata}{\textrm{omd}}
\newcommand{\question}{\rq}                  
\newcommand{\image}{\rv}
\newcommand{\longthought}{\ermZ}
\newcommand{\stepthought}{\rz}

\newcommand{\reasoning}{\rz}                 
\newcommand{\densecaption}{\rc}
\newcommand{\step}{\rs}

\newcommand{\marker}{{\textnormal{m}}}              
\newcommand{\thought}[1]{\tau_{#1}}           
\newcommand{\thoughts}{\boldsymbol{\tau}} 

\newcommand{\newreasoning}{\bar{z}}                 

\newcommand{\questionspace}{\mathcal{Q}}   

\newcommand{\answer}{\ra}                    
\newcommand{\answerspace}{\mathcal{A}}     

\newcommand{\basemodel}{\texttt{BaseModel}\xspace}

%% file: sections/1-intro.tex
\begin{figure*}[t!]
    \centering
    \includegraphics[width=1.01\linewidth]{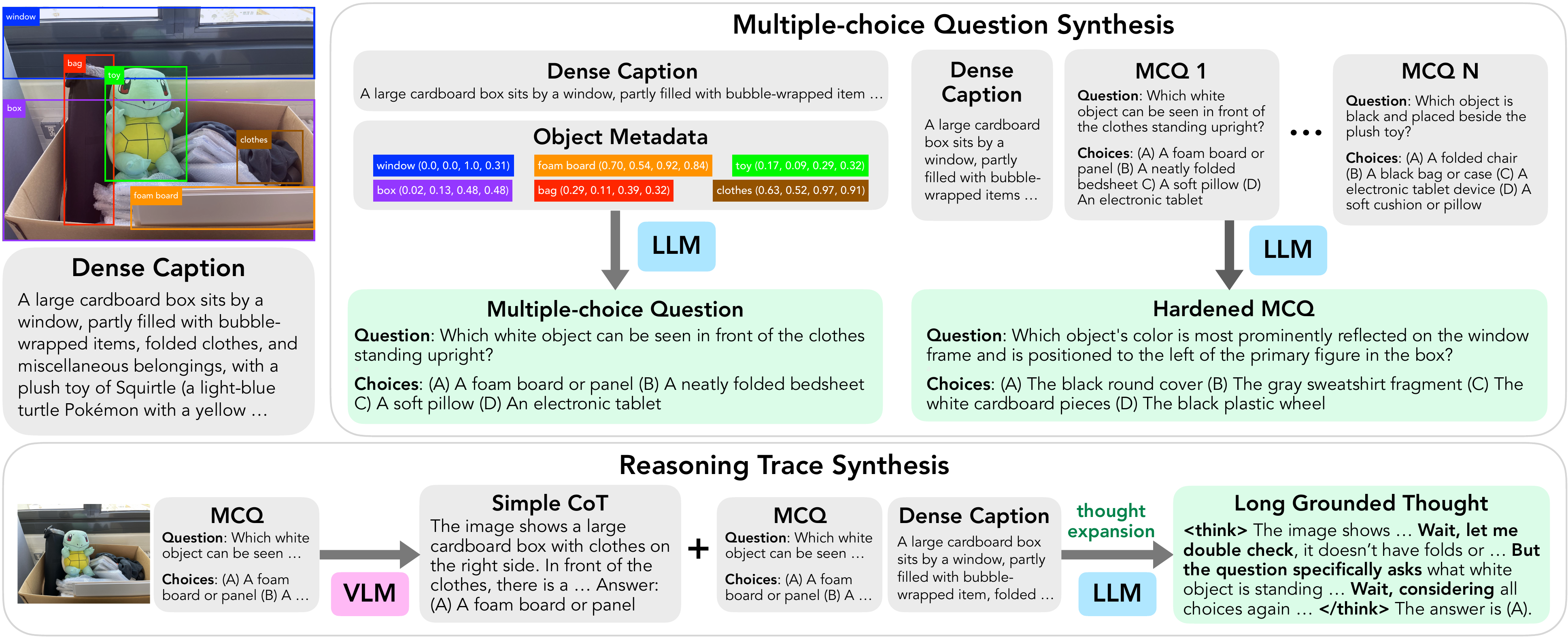}
    \caption{\textbf{Overview of our two-stage \rebuttal{data generation} framework}. First, we synthesize multiple-choice questions (MCQs) from dense captions and grounded object metadata, emphasizing scale and diversity. Later, we harden questions by composing them into  visual reasoning problems that requires decomposition and higher-order cognitive patterns. For each stage, we also synthesize reasoning traces by first distilling CoTs from VLMs and then expanding them with reasoning LLMs, yielding traces that are in the distribution of VLM outputs yet richer in reasoning depth. \rebuttal{The caption is never directly seen by the trained model.}}
    \label{fig:pipeline_main_figure}
    \vspace{-2mm}
\end{figure*}

\section{Introduction}

Since the arrival of DeepSeek R1~\citep{deepseekai2025deepseekr1incentivizingreasoningcapability}, 
a wealth of open-source reasoning datasets has been developed for language-based reasoning~\citep{muennighoff2025s1simpletesttimescaling,jung2025prismatic,openthoughts,lin2025goedelproverv2scalingformaltheorem}, as innovations in data curation and distillation have proven to be powerful levers for advancing open-source models. 
In contrast, open-source multimodal reasoning efforts seem to lag behind. Perhaps because it is not entirely obvious how to synthesize new vision-centric problems with an appropriate level of complexity and  long chain-of-thoughts (CoTs) with complex reasoning structures (e.g., verification, backtracking, subgoal setting) \textit{specifically} for vision-centric reasoning tasks.
Indeed, most open vision-centric reasoning datasets with complex problems and reasoning structures remain either limited in scale or  focused on visual math.
Table~\ref{open_datasets_comparison} highlights this by comparing  popular multimodal reasoning datasets that exhibit traces with complex  structures.

\new{
A recent step forward came from LongPerceptualThought \citep[LPT;][]{liao2025longperceptualthoughts}, which synthesized 30K problems with structured reasoning traces for vision-centric tasks. By using dense captions as a proxy between images and language, LPT combines VLMs with reasoning LLMs to synthesize new problems and traces with complex cognitive structures. However, we identify two critical bottlenecks in their framework when pushed to scale: 1) \textbf{problem synthesis saturation}, relying solely on captions for problem synthesis limits the diversity and grounding of the generated vision-centric questions, causing downstream performance to saturate quickly and 2) \textbf{cognitive simplicity}, the synthesized reasoning traces for those problems—while richer than previous works—still lack higher-order structures.
}

%

%

\new{In this work, we push this line of research further with a synthesis framework that tackles these three core challenges: \textbf{scale}, \textbf{complexity}, and \textbf{CoT richness}. We introduce \ourdataset, containing \textit{1M+ new vision-centric problems}  with complex reasoning traces
distilled from frontier models (e.g., Qwen2.5-VL, R1-671B) 
including \textit{129K additional problems for offline and online RL}}. 

\new{Compared to recent methods like LPT~\citep{liao2025longperceptualthoughts}, our framework delivers significant improvements:
\begin{itemize}
    \item \textbf{Scale:} We overcome the bottlenecks of caption only synthesis methods such as LPT by leveraging grounded object metadata. This allows us to scale problem synthesis from 30K to over 1M+ high-quality vision centric problems without  performance saturation.
    \item \textbf{Complexity:} We significantly increase vision-centric problem synthesis difficulty. Our composition hardening reduces the rate of trivially solvable questions by $\sim$10$\times$ (decreasing from 36.7\% in baselines to 3.3\%).
    \item \textbf{CoT Richness:} We enhance the cognitive depth of the synthesized CoT traces. Our method increases the frequency of complex cognitive behaviors by +206\%, yielding traces that are 3$\times$ richer than the baseline.
\end{itemize}}

We show that finetuning a 7B VLM on our data outperforms all open-data baselines on several vision-centric benchmarks. Notably, our best configurations match or outperform strong closed-data models like MiMO-VL-7B-RL in 3 out of 5 benchmarks.
Even without any video or embodied-QA data in our dataset, our fine-tuned model achieves substantial improvement 
on  an \textit{open-ended} embodied QA task~\citep{kim2025needlesembodiedhaystackenvironment}. Perhaps more surprising, 
despite being entirely vision-centric, our data also transfers positively to text-only reasoning (MMLU-Pro) and audio reasoning (sound and music) on an Omni-7B model. \new{We additionally show that training on our verifiable multiple choice problems also generalize to open ended visual tasks.}

\new{Finally, we leverage our dataset to conduct a comprehensive analysis of the full VLM post-training spectrum for vision-centric tasks. To the best of our knowledge, \textbf{this represents one of the largest studies  of its kind at scale on VLMs to date} revealing four critical insights:} \textbf{(i)} online RL necessitates offline “teaching” of cognitive behaviours—instruct models without cognitive skills underperform SFT on high-quality synthetic data; \textbf{(ii)} multi-staged offline training (SFT→DPO) could reach competitive performance compared to online RL with decoupled compute and higher scalability; \textbf{(iii)} online RL on base models yields early gains but plateaus quickly; 
\textbf{(iv)} high quality SFT alone can substantially improve out-of-domain, cross-modality transfer.



\begin{table*}[t!]
\centering

\caption{Comparison of our visual reasoning dataset with prominent open-source counterparts. Our dataset scales to over 1M+ examples. \rebuttal{ Cognitive behaviours are marked as present (\cmark) if the average count on a sampled subset of 1000 examples is at least 10\%, we quantify the behaviour following the methodology from~\citet{gandhi2025cognitivebehaviorsenableselfimproving,liao2025longperceptualthoughts}.}}
\begin{adjustbox}{width=\linewidth}
\begin{tabular}{@{}lrlccccc@{}}
\toprule
& & & & \multicolumn{3}{c}{\textbf{Cognitive Behaviours}} \\ \cmidrule(l){5-7}
\textbf{Dataset} & \textbf{\# QA} & \textbf{Primary Domain} & \textbf{Data Type} & \textbf{Subgoal} & \textbf{Backtrack} & \textbf{Verify}  \\
\midrule
PixMo-AskModelAnything~\citep{deitke2024molmo} & 162K &  Visual Question Answer&  SFT & \xmark & \xmark & \xmark  \\

%
SCI-Reason~\citep{ma2025scireason} & 12.6k & Scientific Image Reasoning & SFT & \xmark & \xmark & \xmark \\
LENS~\citep{wang2024multi} & 40k & Multi-Scenario Visual Reasoning   & SFT & \xmark & \xmark & \xmark \\
DriveLMM-o1~\citep{ishaq2025drivelmm} & ~22k & Driving Visual Reasoning   & SFT & \cmark & \xmark & \xmark \\
Virgo~\citep{du2025virgo} & ~19K & Visual Math & SFT & \cmark & \cmark & \cmark  \\
VLLA-Thinking~\citep{chen2025sft} & ~152K & Multimodal Reasoning \& Visual Math & SFT/RL & \cmark & \cmark & \cmark  \\
LongPerceptualThoughts~\citep{liao2025longperceptualthoughts} & 30k & Vision-Centric Reasoning & SFT/RL & \xmark & \cmark & \cmark  \\
\midrule
\textbf{Ours} & \textbf{1M+} & \textbf{Vision-Centric Reasoning \& Compositional Reasoning} & \textbf{SFT/RL} & \text{\cmark} & \text{\cmark} & \text{\cmark}  \\
\bottomrule
\end{tabular}%
\end{adjustbox}
\label{open_datasets_comparison}
\end{table*}

%% file: sections/2-method.tex
\vspace{-2mm}
\section{Method}

Our data generation framework, illustrated in Figure~\ref{fig:pipeline_main_figure}, tackles three core challenges: \textbf{scale}, \textbf{complexity}, and \textbf{richness} of the reasoning trace.
In the first stage, we generate large-scale  multiple choice visual problems (MCQs) using high-quality captions and grounded metadata (i.e., bounding boxes).  \new{This stage emphasizes synthesis of diverse problem at  scale
(Section~\ref{method:scale_diversity}). }
In the second stage (Section~\ref{method:complexity}), we apply a composition hardening algorithm that merges MCQs from the first stage into more challenging, multi-hop problems that require decomposition and higher-order reasoning.
For each stage, we then synthesize reasoning traces that could be used for SFT and offline RL (Section~\ref{method:synthesis}). 
In our notation, $\VLM$ denotes the VLM, $\LLM$ the LLM, and $\RLLM$ the reasoning LLM.

\subsection{Scale and Diversity}\label{method:scale_diversity}

We focus on synthesizing visual problems (i.e., multiple-choice questions, MCQs) at scale. Two key requirements for scaling MCQ synthesis are \textbf{simplicity} and \textbf{diversity}. Simplicity ensures that we can generate a large number of MCQs in a cost-efficient manner, while diversity guarantees that each question is unique and non-redundant. A straightforward approach is to provide an LLM with a highly detailed description of the image and prompt it to generate MCQs based on the image and its associated dense descriptions. 
%

\begin{figure}[h!]
    \centering
    \includegraphics[width=\linewidth]{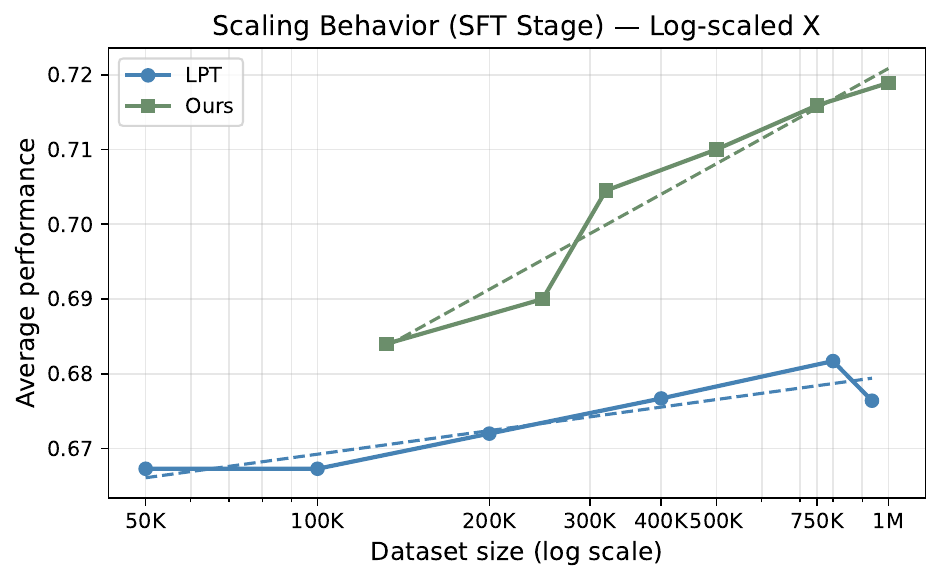}
    \caption{Scaling Behaviour of LPT vs Ours for SFT. We find that using additional metadata (here bounding boxes) in addition to highly details captions allows for more diverse and controlled generation of MCQs, successfully scaling beyond 1M+ examples. \rebuttal{Please refer to appendix~\ref{suppl:diversity_analysis} for additional analysis on diversity.}}
    \label{fig:scaling}
    \vspace{-3mm}
\end{figure}

%
%
\new{
However, scaling this caption-only approach to millions of examples reveals a critical bottleneck: question diversity quickly saturates, causing downstream performance to plateau (Figure~\ref{fig:scaling}). We attribute this to \textit{problem synthesis saturation}, where the synthesizer (LLM) repeatedly targets the same salient objects on the caption and generate similar questions about similar objects. Our embedding analysis (Appendix~\ref{suppl:diversity_analysis}) confirms that grounding on objects mitigate this. Compared to the baseline, our method yields a \textbf{3.2$\times$ wider semantic spread} and reduces redundancy (average pairwise cosine similarity of \textbf{0.61} vs. 0.82). Most importantly, Figure~\ref{fig:scaling} shows \textbf{positive downstream performance} and slope even after scaling to  1M+ visual problems. 
}

Formally, given an \textbf{image} $\image$ with dense textual \textbf{descriptions} $\densecaption$, and object-level \textbf{metadata} $\objectmetadata$ (bounding-box coordinates and object tags), our goal is to construct a triplet:
\[
(\image, \question, \answer) := \LLM( \densecaption, \objectmetadata),
\]
where $\question$ is the generated question and $\answer$ is the correct answer. 
Intuitively, for an image with $K$ detected objects, we can produce $K$ distinct MCQs by conditioning on each object. Interestingly, we found that including normalized bounding-box coordinates further improved question grounding, even though the LLM itself operates purely in   text .  
Finally, we apply filtering protocol to ensure quality.
\rebuttal{ See Stage 1~(\ref{suppl:stage_1} for details).} 
%
%
We assume object metadata is not given and first processed images with Grounded-SAM~\citep{ren2024grounded}, producing open-vocabulary bounding boxes and object tags.  
%

%

\begin{figure*}[t]
    \centering
    \begin{subfigure}[b]{0.48\linewidth}
        \centering
        \includegraphics[width=\linewidth]{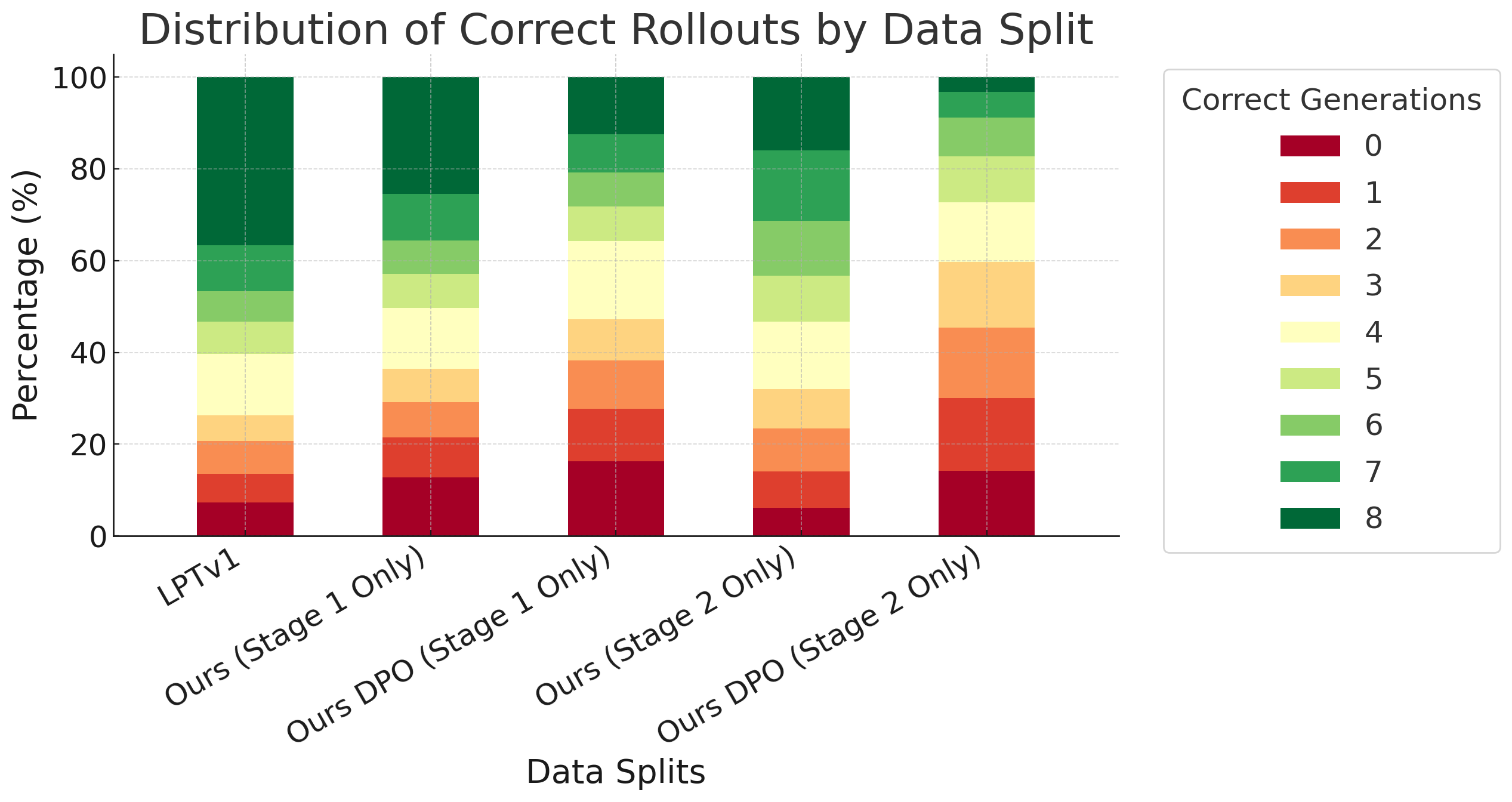}
        \caption{Complexity estimation via multiple rollouts. 
        }
        \label{fig:complexity}
    \end{subfigure}
    \hfill
    \begin{subfigure}[b]{0.48\linewidth}
        \centering
        \includegraphics[width=\linewidth]{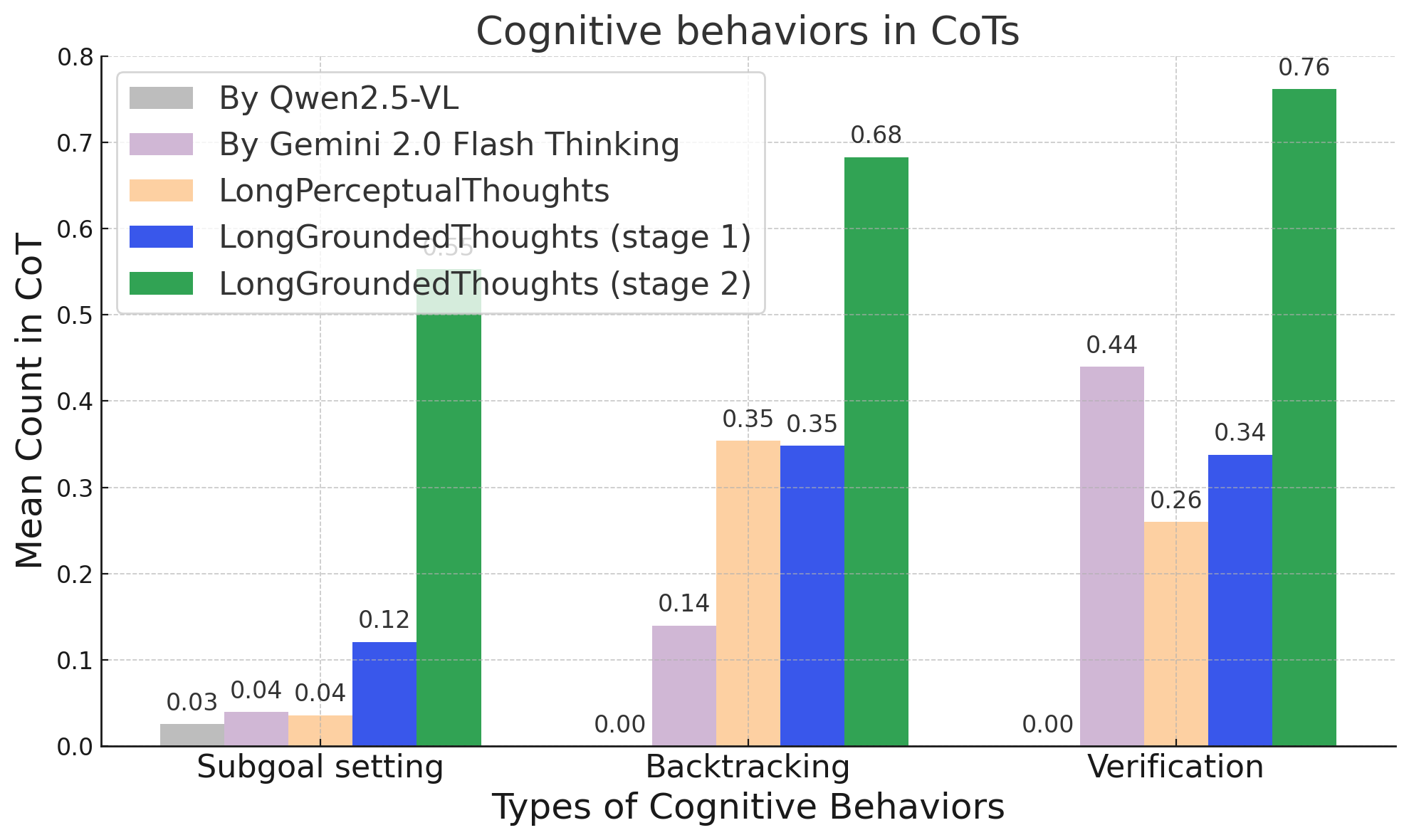}
        \caption{Analysis of Cognitive Behaviors in CoTs. 
        }
        \label{fig:cogitive}
    \end{subfigure}
    \caption{\textbf{Analysis of our data splits.} \textbf{(a)}~Complexity estimation via multiple rollouts on synthesized MCQs using Qwen2.5-VL as a policy. Darker green color represents \textit{easier} problems. \textbf{(b)}~Analysis of Cognitive Behaviors in CoTs. Our data exhibits higher frequencies of subgoal setting, backtracking, and verification, indicating a more deliberate and structured reasoning process. Estimation of cognitive behaviours  and terminology from ~\cite{gandhi2025cognitivebehaviorsenableselfimproving}.
    \rebuttal{Table~\ref{tab:appendix-stats-detailed-behaviours} shows detailed cognitive behaviours analysis on several  datasets.}
    }
    \label{fig:combined}
\end{figure*}

\begin{figure*}[t]
    \centering
    \includegraphics[clip,width=0.92\linewidth]{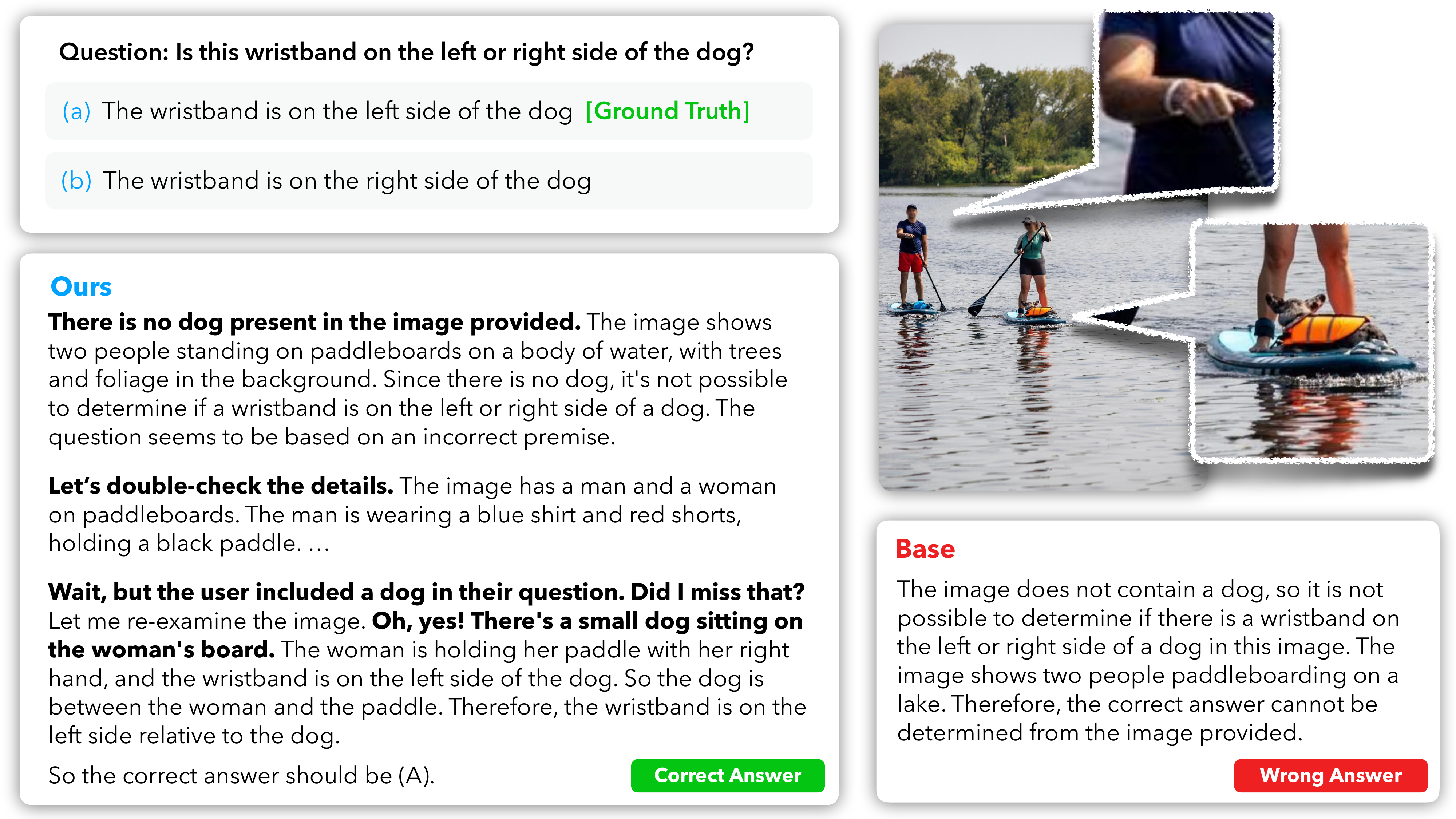}
    \caption{
    \textbf{\rebuttal{Reasoning trace comparison between our model (post-SFT and RL) and the vanilla base model.}}
    \rebuttal{Both models initially fail to identify the dog in the image. The base model terminates with an incorrect answer based on this flawed premise. In contrast, the model finetuned on \ourdataset demonstrates a non-linear reasoning process; it employs self-verification and backtracking to challenge and self-correct its initial assessment. This correction appears to stem by relying on captioning and grounding as a bridge between language and vision; notably  grounding on the dog triggers the revised  path on a second 
    verification structure. 
    }}
    \label{fig:example_dog}
    \vspace{-3mm}
\end{figure*}

\subsection{Compositionality for Complexity}\label{method:complexity}
A limitation of synthetic MCQ generation 
is that the resulting questions are often relatively easy. In practice, we observed a large percentage of these problems can be solved directly by the base VLMs. As shown in Figure~\ref{fig:complexity} and Table~\ref{tab:complexity_richness}, our simple \new{grounded synthesis strategy (Stage 1)} already generates problems that are harder than caption-only methods, yet many remain solvable for the base model. 

To push synthesis beyond this regime, we design a simple, scalable composition algorithm that leverages an LLM to merge multiple questions into a harder one. Specifically, the algorithm selects $K$ MCQs associated with the same image and composes them into a single, more complex problem, where the original MCQs act as intermediate steps.  
Formally, given an \textbf{image} $\image$ with dense textual \textbf{descriptions} $\densecaption$, a collection of questions and answers  obtained from the previous stage,  our goal is to construct a composed triplet:
\[
(\image, \question^\star, \answer^\star) := \LLM(\densecaption, \{\question_i, \answer_i\}_{i=1}^K),
\]
where $\question^\star$ is the synthesized complex question and $\answer^\star$ its corresponding solution.  
\rebuttal{(see~\ref{appendix:a:3} for prompts and examples)}. 
We refer to this data as Stage 2.
\rebuttal{We define multi-step reasoning as a reasoning trace that decomposes the original problem into manageable sub-steps to reach the final answer (i.e. subgoal setting). As shown in Figure~\ref{fig:cogitive} stage 2 leads to a significantly higher subgoal setting.
Figure~\ref{fig:complexity}, on the other hand shows the generated questions are notably harder. 
}

\vspace{-2mm}
\subsection{Synthesizing Reasoning Traces}
\label{method:synthesis}
   We synthesize long CoTs,  
using the two stage expansion strategy presented in LPT~\citep{liao2025longperceptualthoughts}. 
We prompt a VLM $\VLM$ with the image and its corresponding MCQ to produce a rationale ($\stepthought_1$) and final prediction ($\answer_1$), denoted as $(\stepthought_1, \answer_1) := \VLM(\image, \question)$. 
%
Then, we structure a prompt as:
\begin{align*}
\text{User: }  \densecaption \oplus \question ,
\text{Assistant: }  \texttt{<think>} \oplus \stepthought_1
\end{align*}
and ask $\RLLM$ to continue the thought, producing $(\stepthought_2, \answer_2)$, here $\oplus$ denotes concatenation. 
Figure~\ref{fig:cogitive} analyzes cognitive behaviors in our reasoning traces, highlighting notable non-linear patterns in our data, especially in hardened MCQs from stage 2.
\rebuttal{We emphasize the caption is never seen for the training model or during inference. The caption is used only by the reasoning LLM during the expansion as the LLM cannot "see". 
}

\new{
\textbf{Do we need expanded long CoTs?} Yes. We ablated the long CoT expansion strategy on 750K examples (Table~\ref{tab:cot_ablation}). Surprisingly, training on simplified `Short CoT` distilled only from VLM traces degrades performance below even `No CoT` (direct answers), likely due to negative transfer from low-fidelity reasoning. Only `Long CoT' traces obtained by sampling from the VLM and expanded with complex cognitive behaviors via the reasoning LLM yields consistent gains, validating the necessity of the expansion.
}

\textbf{Scaling challenge.} At larger scales, we observed that $\RLLM$ frequently referenced the provided image description explicitly (e.g., “the image description says,…”), rather than producing an expansion as if grounded in the visual input. 
To overcome this, we introduce a guided decoding mechanism on the sampling algorithm that uses regular expressions (regex) to constrain generations.
This substantially improves distillation efficiency and utilization at scale.

\textbf{Multi-Stage Synthesis.} 
For MCQs generated in Stage~1 (simpler problems), we use Qwen2.5-VL-7B for the initial VLM synthesis and DeepSeek-R1-Distill-Qwen-32B as our $\RLLM$ for expansion. For Stage~2 (harder, composed problems), we reserve the strongest available models—Qwen2.5-VL-72B, R1-671B and Qwen3-235B-Thinking for reasoning LLM expansion.  
\new{Crucially, 75\% of our data uses accessible models (Stage 1) yet yields strong performance (Table~\ref{tab:rl_sft_ablation}). We release both subsets separately to enable both reproducibility and high-quality research without  dependence on frontier teachers.}

\subsection{Offline and Online Synthesis for RL}
To build a \textit{preference dataset} for offline RL, we follow~\citep{setlur2024rlincorrectsyntheticdata,zhang2025backtracking,kimiteam2025kimik15scalingreinforcement} and define pairwise comparisons based on \emph{correctness} and \emph{compactness}. 
%
Formally, we define:
\begin{align*}
\text{Correctness: } & (\stepthought_1^+, \answer_1^+) \succ (\stepthought_1^-, \answer_1^-),  \\
& (\stepthought_1^- \oplus \stepthought_2^+, \answer_2^+) \succ (\stepthought_1^-, \answer_1^-) \\
\text{Compactness: } & (\stepthought_1^+, \answer_1^+) \succ (\stepthought_1^+ \oplus \stepthought_2^+, \answer_2^+).
\end{align*}
Here, the superscript $+$ denotes a correct prediction and $-$ an incorrect one, while $\succ$ indicates preference and $\oplus$ denotes concatenation. Following previous works, 
%
We specifically collect both positive and negative CoTs from the initial VLM response (e.g., $(\stepthought_1^+, \answer_1^+)$ and $(\stepthought_1^-, \answer_1^-)$), together with subsequent reasoning expansions from $\RLLM$ (e.g., $(\stepthought_2^+, \answer_2^+)$). 
%
Note that we do not explicitly verify whether $\stepthought_1$ itself is correct or incorrect; instead, we infer its correctness from the associated $\answer_1$, which we find to be a reasonable approximation at scale.   
\new{For \textit{online RL} (GRPO), we only keep the generated visual problems and its verifiable answer. }

%% file: sections/3-exp.tex
\begin{table*}[t]
\centering
\footnotesize
\setlength{\tabcolsep}{5pt}
\renewcommand{\arraystretch}{1.12}
\caption{\textbf{Main results on vision-centric reasoning benchmarks.} We compare our models against both open- and closed-source VLMs across \textit{five} challenging benchmarks. 
Community baselines trained on open source data sometimes underperform the base Qwen2.5-VL-7B-Instruct, underscoring the lack of high-quality open reasoning data for \textit{vision-centric tasks}.
\rebuttal{Closed source performance  is shown for reference, taken from the authors technical reports. }
}
\begin{adjustbox}{width=\linewidth}
\begin{tabular}{l cc r r r r r}
\toprule
\textbf{Model} & \textbf{Open Model} & \textbf{Open Data} & \textbf{V* Bench} & \textbf{CV Bench} & \textbf{MMVP} & \textbf{RealWorldQA} & \textbf{MMStar-V} \\
\midrule
Qwen2.5-VL-7B-Instruct & \cmark & \xmark & 75.39 & 74.52 & 74.67 & 67.84 & 65.60 \\
MiMo-VL-7B-SFT & \cmark & \xmark & 80.60 & 81.80 & \textbf{78.33} & 71.90 & 67.60 \\
MiMo-VL-7B-RL  & \cmark & \xmark & 81.70 & 82.30 & 77.67 & \textbf{72.68} & 67.07 \\
\midrule
GPT-4o  & \xmark & \xmark & 73.90 & 76.00 & \textemdash  & \textemdash  & \textemdash   \\
o1 & \xmark & \xmark & 69.70 & \textemdash & \textemdash  & \textemdash  & \textemdash \\
Claude 3.7 & \xmark & \xmark &  \textemdash & 75.40 &  \textemdash & \textemdash   & \textemdash   \\

\midrule
Qwen2.5-VL-7B-Instruct &   &  &  &   &   &   \\
+ VLAA-Thinker  & \cmark & \cmark & 56.54 & 72.95 & 75.00 & 66.93 & 63.07 \\
+ Revisual-R1-final & \cmark & \cmark & 68.58 & 72.77 & 65.33 & 62.48 & 60.80 \\
+ LongPerceptualThoughts   & \cmark & \cmark &  80.60 & 75.30 & 77.00 & 67.45 & 64.13 \\
\midrule
+ Ours (SFT) & \cmark & \cmark & 79.05 & 80.60 & 73.67 & 65.49 & 64.27 \\
+ Ours (SFT + DPO) & \cmark & \cmark & 80.10 & 81.51 & 74.00 & 66.14 & 64.40 \\
+ Ours (Multistage SFT + DPO) & \cmark & \cmark & \textbf{83.25} & 82.28 & 72.33 & 68.76 & 66.27 \\
+ Ours (SFT + GRPO) & \cmark & \cmark & 81.68 & \textbf{83.80} & 72.00 & 69.02 & \textbf{68.40} \\
\bottomrule
\end{tabular}
\end{adjustbox}
\label{tab:main_results}
\end{table*}

\begin{table}
\centering
\small
\setlength{\tabcolsep}{4pt}
\caption{\textbf{Quantitative comparison of Data Complexity and CoT Richness.} We compare our Stage 1 and Stage 2 data against the LPT baseline. Lower Pass/Solve rates indicate higher complexity; higher Behavior counts indicate richer reasoning traces.}
\label{tab:complexity_richness}
\begin{tabular}{@{}lccc@{}}
\toprule
\textbf{Method} & \textbf{Pass Rate} $\downarrow$ & \textbf{Perfect Solve} $\downarrow$ & \textbf{CoT Richness} $\uparrow$ \\
& \scriptsize{(Avg. over 8)} & \scriptsize{(8/8 Correct)} & \scriptsize{(Behaviors/Trace)} \\
\midrule
LPT (Baseline) & 66.1\% & 36.7\% & 0.65 \\
Ours (Stage 1) & 56.8\% & 25.5\% & 0.80 \\
\textbf{Ours (Stage 2)} & \textbf{38.4\%} & \textbf{3.3\%} & \textbf{1.99} \\
\bottomrule
\vspace{-10mm}
\end{tabular}
\end{table}

\vspace{-3mm}

\section{Experiments}
\new{In this section, we first compare our synthesized dataset vs a SoTA baseline (LPT) in terms of scale, downstream performance, complexity of the synthesized visual problems, and richness of the distilled CoT trace. Later,} we  evaluate the performance of finetuning a  7B VLM on our data, and compare it  compare against three model/data categories: (i) open-weight, open-data models, (ii) open-weight, closed-data models, and (iii) fully closed models.
Next, we use our data to systematically analyze the different post-training stages in a base instruction model, revealing consistent findings.
Finally, we evaluate whether our vision-centric reasoning data transfers to a different modality and architecture (e.g. text-only reasoning, embodied QA, and audio reasoning on an Omni-7B model). \new{We also show that the model trained on \ourdataset generalizes to open-ended visual  problems.}
%

\vspace{-2mm}
\subsection{Scale, Complexity and Richness of CoT}
\new{We quantify improvements over the LPT baseline across three key dimensions:
\textbf{Scale.} Figure~\ref{fig:scaling} illustrates SFT scaling under identical training conditions. Unlike the baseline, our framework successfully scales beyond 1M+ visual problems while maintaining positive slope. \textbf{Complexity.} We measure complexity via \textbf{Pass Rate} consistency over 8 rollouts (Table~\ref{tab:complexity_richness}, Figure~\ref{fig:combined}). LPT exhibits a high perfect solve rate of \textbf{36.7\%}, indicating many trivially solvable questions. Our data reduces this to \textbf{3.3\%} ($\sim$10$\times$ reduction) and lowers the average pass rate (66.1\% $\to$ 38.4\%), confirming that composition hardening shifts the distribution toward significantly more complex problems.
\textbf{CoT Richness.} We quantify richness by the average count of distinct cognitive behaviors per trace. LPT averages only 0.65 behaviors; our Stage 2 data averages \textbf{1.99} (\textbf{+206\%}), yielding traces that are structurally \textbf{$\sim$3$\times$ richer}.
}

%
\subsection{Setup and Main Results}
\textbf{Base Model and Benchmarks}  
We fine-tune Qwen2.5-VL-7B-Instruct~\citep{bai2025qwen25vltechnicalreport} and
 evaluate on vision-centric tasks.
Following~\cite{liao2025longperceptualthoughts}, for general-knowledge datasets, we retain only their vision splits (e.g., MMStar-V~\citep{chen2024we}). In addition, we adopt vision-centric focused benchmarks   V$^{*}$ Bench~\citep{v_star}, CV-Bench (averaged both 2D and 3D),  MMVP~\citep{tong2024cambrian} and RealWorldQA~\citep{RealWorldQA_xAI_2024}. These benchmarks test visual search, 2D/3D spatial reasoning, fine-grained attribution, and  scene understanding. 

\textbf{Evaluation.} We use two main protocols for evaluation. For comparison vs existing models and published results we used VLMEvalKit~\citep{duan2024vlmevalkit} with GPT-4o as a judge (Table~\ref{tab:main_results}).
For ablations, we utilize the evaluation protocol of ~\citep{liao2025longperceptualthoughts} which utilizes rule-based matching instead of a judge. For clarity, results are shown with decimal when no LLM as a judge is used. 

\textbf{Source of high-quality image captions and metadata.}  
We use DOCCI~\citep{OnoeDocci2024}, a human-annotated dense caption dataset.
We defer exploring other caption datasets and automatically generated captions to future work.

\textbf{Baselines.}  
We compare against three categories: (i) open-weight, open-data models, (ii) open-weight, closed-data models, and (iii) fully closed models.  For \textit{open-weight, open-data}, we include VLAA-Thinker~\citep{chen2025sft}, a model trained on $\sim$152K synthetic reasoning traces
using both SFT and RL. We also compare with LPT~\citep{liao2025longperceptualthoughts} and  ReVisual-R1~\citep{chen2025advancingmultimodalreasoningoptimized}.
Additionally, we include Virgo~\citep{du2025virgo}, trained on $\sim$19K math-heavy datapoints.   For \textit{open-weight, closed-data}, we consider MiMo-VL-7B, a SoTA VLM trained with a four-stage recipe totaling $\sim$2.4T tokens across images, videos, and text. 
\rebuttal{For \textit{fully closed models}, we report results for GPT-4o and Claude~3.7 
taken from~\citep{xiaomi2025mimo}.
}

\textbf{Training Algorithms.}

\textbf{SFT.}  
We perform SFT on the large pool of data from Stage~1 and Stage~2. To simplify training, we adopt a curriculum: first fine-tuning on simpler Stage~1 data for one epoch, then introducing the more complex Stage~2 data. Stage~1 consists of $\sim$750K datapoints, and Stage~2 adds $\sim$251K. We fine-tune the language decoder with a batch size of $256$,  learning rate $ 8 \times 10^{-6}$. For Stage~1, we train for up to one epoch with maximum image resolution $512 \times 512$ and input cutoff length $1024$. For Stage~2, we halve the learning rate and apply early stopping based on validation loss. All experiments are implemented with \texttt{llama-factory}.  

\textbf{Offline RL (DPO).}  
We fine-tune the language decoder with a batch size of $256$ using a  learning rates of $\{6 \times 10^{-6}\}$. Training runs for up to six epochs with early stopping based on validation loss. For DPO, we set $\beta = 1$ and, following~\cite{pang2024iterative}, include the SFT loss with a weight of $0.5$. Implemented with \texttt{llama-factory}.  

\textbf{Online RL (GRPO).}  
We perform GRPO on top of the fine-tuned model via \texttt{VERL} \citep{verl}. We set the batch size 128, max response length 8192 and the learning rate $1 \times 10^{-6}$. We use the KL loss coefficient of $0.001$ and omit entropy penalty. To avoid overfitting to a specific answer format, our labels are set in a diverse format (\textit{e.g.,} (A), A, and (A) \textit{answer}); thus, we compute the reward based on an LLM judge, assigning reward = $1$ to a correct answer and reward = $0$ to a wrong answer (as identified by the judge). We additionally assign format reward, we add $0.1$ to the reward if the  response is formatted correct (\textit{i.e.,} the model correctly generates both \texttt{<think>} and \texttt{</think>}).

\textbf{Main Result. Comparison vs. Baselines.}  
Table~\ref{tab:main_results} compares Qwen2.5-VL-7B fine-tuned with our data against community baselines, close data, and closed models. Models trained on limited open reasoning data (e.g., VLAA-Thinker, Revisual) often underperform even the base Qwen2.5-VL-7B-Instruct, underscoring the lack of high-quality open data for vision-centric reasoning. In contrast, our models consistently improve the base, achieving the best results and outperforming on 3/5 benchmarks.

\new{
\textbf{Generalization to open-ended non-MCQ tasks.} Despite leveraging entirely multiple-choice questions, our models generalize robustly to open-ended visual problems and unseen domains. Table~\ref{tbl:main_results_combined} shows we achieve a \textbf{+8.8 point gain} on NiEH~\citep{kim2025needlesembodiedhaystackenvironment}, an \textbf{open-ended} embodied benchmark. We also observe consistent improvements on MathVision (mixed format also including \textbf{open-ended} tasks) (see Table~\ref{tab:supp_math_visulogic}). This confirms that our reasoning gains transfer beyond the training format and domain.
}
\begin{figure*}[t!]
    \centering
    \includegraphics[clip,width=0.92\linewidth]{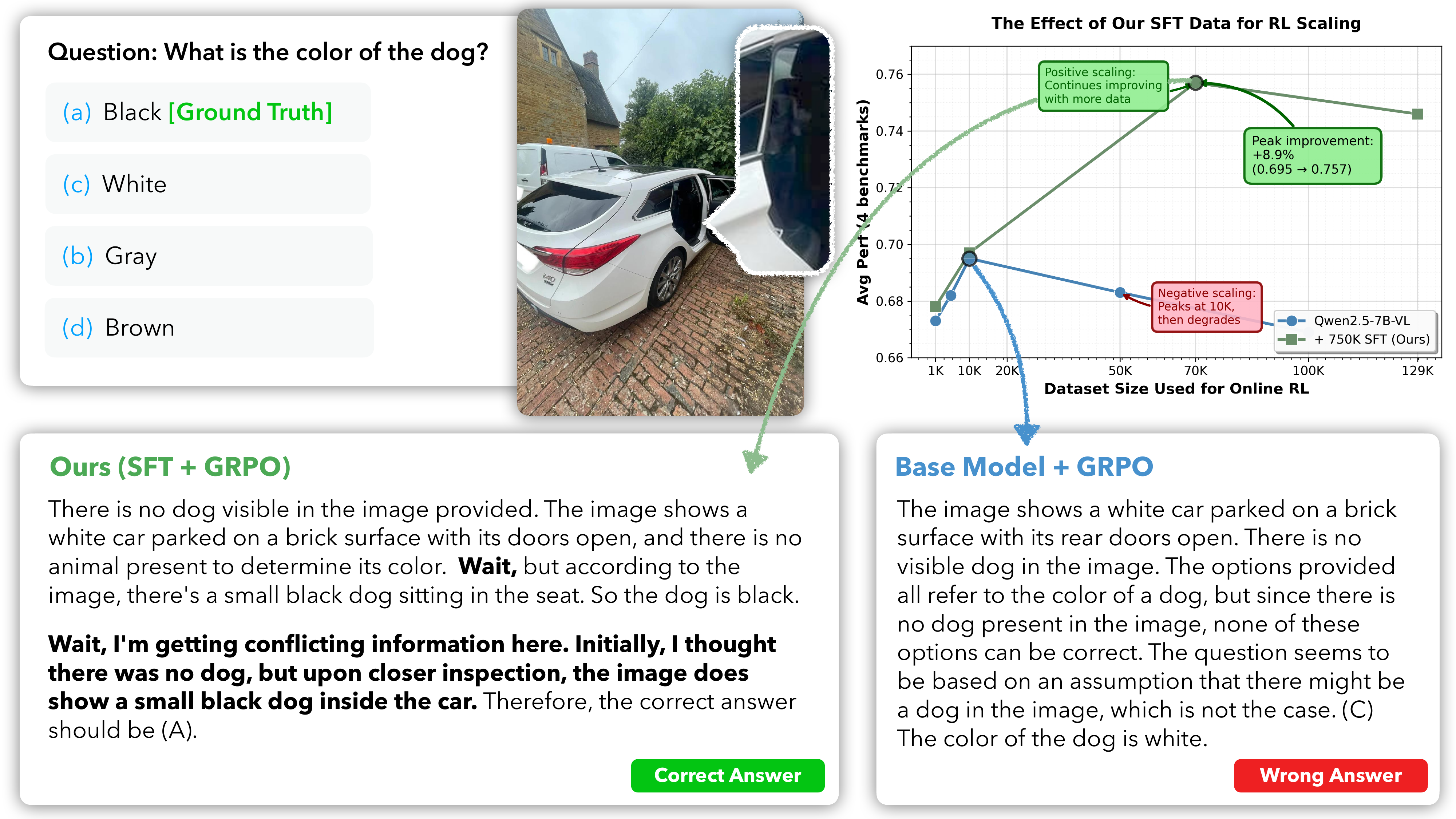}
  \caption{
    \textbf{\rebuttal{Quantitative and qualitative comparison of the post-training pipeline on our data  vs pure RL on the base model}}.
    \rebuttal{
     \textbf{(Upper Right)} The graph illustrates the effect of scaling dataset size during online RL showcasing relative peak improvement in percentage. The baseline (blue line), starting from an off-the-shelf model, exhibits \textit{negative scaling}: performance peaks at 0.695 (10K samples) and degrades with more data. In contrast, our method (green line), which includes SFT on our high-quality data with complex reasoning traces, allows to scale online RL further. This suggests that without offline "teaching" of reasoning patterns via SFT, online RL fails to effectively utilize larger datasets.
    \textbf{(Upper Left)} A qualitative example (from V* bench), using each model's best checkpoint (indicated by a dot on the curve), highlights the resulting difference in reasoning. The baseline model fails to identify the partially obscured dog and answers incorrectly. Our model also initially expresses confusion but then self-corrects 
    ("Wait, I'm getting conflicting information...")
    , showcasing a multi-step reasoning process to arrive at the correct answer. This self-correction capability, instilled with our data, is not observed in the baseline, indicating RL alone was insufficient to elicit this behavior.
    Image brightness was increased for illustration purposes.
    }
  }
    \label{fig:qual_quant_ablation}
    \vspace{-4mm}
\end{figure*}

\subsection{Analysis of VLM Post-training Stages}
We analyze VLM post-training strategies using our data. Table~\ref{tab:rl_sft_ablation} reports aggregated performance, averaged over four benchmarks. Figure~\ref{fig:scaling} shows scaling compared to LPT. For evaluation, we use rule-based matching instead of LLM-as-judge for cost efficiency. Here, we summarize our findings:

\textbf{Online RL requires prior teaching of cognitive behaviours; starting from the base instruct model underperforms SFT only on our data.}
When starting from  the base model lacking cognitive patterns~(see Figures~\ref{fig:qual_quant_ablation},~\ref{fig:cogitive}), GRPO peaks at \textbf{0.695} (10K) and declines with more data (50–100K), remaining below our SFT baseline (\textbf{0.716}). 
Finetuning first on LPT data improves   but still  underperforms simple SFT finetuning on our data. Finetuning on our data leads to the best results suggesting that without offline “teaching” of cognitive behaviours,  online RL cannot elicit competitive performance vs SFT in high-quality and diverse data. Notably,  we do not observed online RL was able to elicit structured complex reasoning traces unless it was pretrained first on our reasoning SFT data (see Figure~\ref{fig:qual_quant_ablation} for a qualitative example).

\textbf{Offline (staged SFT$\rightarrow$DPO) is competitive to online RL  while decoupling compute needs.}
Our best \emph{offline} configuration (\textbf{SFT 750K + DPO 129K}) attains \textbf{0.740}, within \textbf{1.7} points of the best \emph{online} configuration (\textbf{SFT 750K + GRPO 70K} at \textbf{0.757}). Thus, staged offline preference learning could reach similar ballpark accuracy with no need for synchronized RL compute, making data/compute scheduling more flexible. We additionally observe DPO continue scaling with more data although at a lower rate.

\textbf{GRPO yields fast gains but saturates; scaling past $\sim$70K does not help.}
On top of our SFT stage, GRPO at \textbf{70K} achieves the overall best \textbf{0.757}; increasing to \textbf{129K} drops to \textbf{0.746}. From the base instruct model, performance peaks at \textbf{10K} and degrades thereafter. This is a classic fast-gain/plateau pattern for online RL. Going beyond this regime is still an open question.

\textbf{SFT data diversity matters; grounded MCQ SFT $>$ LPT SFT.}
Our grounded-MCQ SFT reaches \textbf{0.716} versus \textbf{0.682} for LPT SFT (both at comparable scale), a gain of \textbf{+3.4} points on average. 
\subsection{Beyond Vision-Centric Reasoning}

%
\begin{table}
\centering
\caption{\textbf{Out-of-domain evaluation.} We assess cross-domain transfer on MMLU-Pro (text-only) and NiEH (embodied QA). While baselines like Virgo and VLAA degrade text reasoning (negative transfer), our method preserves performance. Additionally, our model generalizes effectively to the open-ended single evidence NiEH benchmark, achieving a \textbf{+8.8 point gain} over the base model despite using only  MCQ training data.}
\label{tbl:main_results_combined}
\begin{tabular}{lcc}
\toprule
\textbf{Model} & \textbf{MMLU-Pro } & \textbf{NiEH EQA } \\
\midrule
Qwen2.5-VL       & 47.15          & 47.55          \\
\midrule
VLAA-thinking    & 21.56          & 47.85          \\
Virgo            & 37.95          & ---            \\
LPT              & 50.77 & 51.95          \\
\midrule
Ours (SFT)       & \textbf{50.82}          & 48.24          \\
Ours (SFT+GRPO)  & 47.00          & 39.10          \\
Ours (SFT+DPO)   & 47.07          & \textbf{56.34} \\
\bottomrule
\vspace{-7mm}
\end{tabular}
\end{table}

\begin{table}
  \centering
  \caption{\textbf{Omni-modality out-of-distribution results on audio MMAU Benchmark(Sound, Music, Speech) and text-only (MMLU-Pro)}. We use Qwen2.5-Omni-7B as baseline.  }
   \begin{adjustbox}{width=\linewidth}
  \begin{tabular}{@{}lccc|c@{}}
    \toprule
    \textbf{Model} & \textbf{MMAU-Sound} & \textbf{MMAU-Music} & \textbf{MMAU-Speech} & \textbf{MMLU-Pro} \\
    \midrule
    Baseline Omni Model                 & 76.77 & 67.33 & 68.90 & 47.00 \\
    \midrule
    + Virgo                    & 64.20 & 59.30 & 64.30          & 39.21 \\
    + LPT                      & 76.63 & 67.56 & 66.93          & 48.74 \\
    
    \midrule

    + Ours (SFT Only)          & \textbf{78.30} & 70.20 & 68.80 & 49.82 \\
    + Ours (SFT + DPO)  & 77.75     & \textbf{70.35}     & \textbf{69.23}              & \textbf{51.07} \\
    \bottomrule
    
  \end{tabular}
  \end{adjustbox}
  \vspace{-4mm}
  \label{tbl:beyond_small}
  
\end{table}

\new{
Table~\ref{tbl:main_results_combined}
shows our model improves upon the base model and existing baselines on text-only reasoning MMLU-PRO~\citep{wang2024mmlupro}, proving our data does not harm text-only capabilities.
Further, despite no video/embodied training, we achieve $\sim$ 8-point gains on the NiEH single-evidence. \textbf{Notably, this is also an open-ended VQA} tasks~\citep{kim2025needlesembodiedhaystackenvironment}. 
Setup details are in Appendix~\ref{app:embodied_section}.
Table~\ref{tbl:beyond_small}, on the other hand, shows that fine-tuning a Qwen2.5-Omni-7B on our data yields consistent cross-modal gains: +0.98 (Sound), +3.02 (Music), and +0.33 (Speech) on MMAU~\citep{sakshi2024mmau}, plus +4.07 on MMLU-Pro. The improvement on MMLU-Pro is more notable in the Omni-model than in the base Qwen2.5-VL. These results demonstrate robust positive transfer from vision-centric data to audio and text modalities.
}

%% file: sections/related.tex
\vspace{-2mm}
\section{Related Work}
Recent efforts have been made to enhance visual reasoning in VLMs, spanning from test-time techniques~\citep{liao-etal-2024-reasoning,liao2024feedbackenhancesemanticgrounding,acuna2025socratic} to improved architectural designs and training pipelines~\citep{cheng2024spatialrgpt,wu2025spatial,chen2025eagle25}. 
However, most efforts to distill high-quality reasoning datasets  at scale remain largely in the text-only domain~\citep{openthoughts,jung2025prismatic,lu2026golden,kim2026privasis}. 
In contrast,
efforts to build 
multimodal datasets 
remain largely linear and/or utilize existing questions rather than synthesizing new problems with different levels of complexity covering the entire post-training VLM pipeline at scale~\cite{xullavacot,ma2025scireason,drivingvqa2025,ishaq2025drivelmm}.
%
A smaller body of work has begun studying more complex reasoning structures such as reflection, self-correction, or iterative refinement in VLMs. For example, Virgo~\citep{du2025virgo} distills multimodal CoTs from multimodal reasoning models~\citep{qvq2024}, focusing on math-heavy domains. VLAA-Thinking~\citep{chen2025sft} similarly distills CoTs in the general domain, though subsequent SFT on these traces has been shown to degrade performance. LPT~\citep{liao2025longperceptualthoughts} introduced long-form CoTs distilled from large reasoning models, such as R1~\citep{deepseekai2025deepseekr1incentivizingreasoningcapability}, with explicit emphasis on expanding thoughts and encouraging models to deliberate more thoroughly across perception-heavy tasks. Our work builds on this line. However, rather than restricting CoTs to fixed stages, linear narratives or reusing existing visual questions, we synthesize diverse and verifiable visual problems with different levels of complexity and complex reasoning traces, scaling along three axes: scale, complexity, and reasoning depth.

%% file: sections/4-discussion.tex
\vspace{-2mm}
\section{Conclusion}
We introduce a framework for synthesizing vision-centric reasoning data that focuses on scale, complexity, and cognitive richness, yielding over 1M high-quality examples. By leveraging grounded metadata and composition hardening, our pipeline generates diverse, verifiable visual problems and complex reasoning structures that significantly surpass prior efforts. Our experiments demonstrate that this data not only improves downstream performance on vision-centric benchmarks but also drives robust positive transfer across modalities and 
out-of-distribution
tasks.
%

%% file: sections/requirements.tex
\section*{Impact Statements}

Our work  contributes to the democratization of visual reasoning capabilities through open datasets and reproducible training recipes. Since our data is synthesized from frontier models, potential risks include the propagation of existing biases found in the teacher models. We do not foresee immediate negative societal consequences beyond the general dual-use concerns applicable to any advancement in visual understanding/reasoning systems.

%% file: sections/appendix_simplified.tex
\section{{Details on Data Generation Stages and Prompts}}

\subsection{{Stage 1: Object-centric Problem Generation}}
\label{suppl:stage_1}
\label{ap:s:1:method}

\begin{figure}[ht!]
    \centering
    \includegraphics[width=0.94\linewidth]{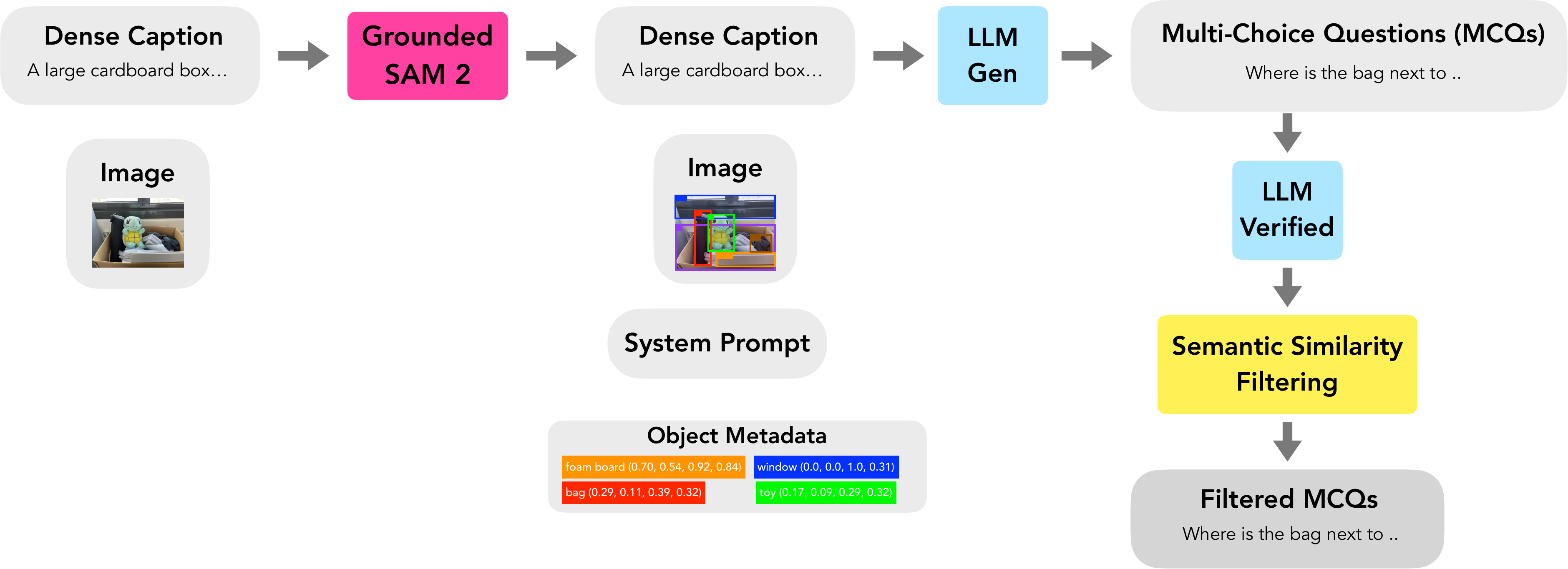}
    \caption{\textbf{Stage 1 Generation Pipeline.}}
    \label{fig:objpt}
\end{figure}

\noindent \textbf{Object-oriented metadata generation.} Our data generation pipeline is composed of two primary stages: a scalable data generation stage focused on quantity and a compositional hardening stage focused on complexity. The first stage is illustrated in Figure~\ref{fig:objpt}.  Our pipeline begins by leveraging a dense captioning model to produce a detailed textual description of the image. Concurrently, we process the image with Grounded SAM-2~\citep{ren2024grounded} to extract a rich set of open-vocabulary bounding boxes and corresponding object tags, which we term Object Metadata. This metadata, including labels and coordinates (e.g., bag ($0.29, 0.11, 0.39, 0.32$)), is integrated with the dense caption into a structured system prompt. This prompt guides a large language model (LLM Gen) to create object-centric MCQs that are inherently focused on specific visual elements. On average, each DOCCI image contains a median of 10.7 bounding tags, determined after applying a confidence cutoff of 0.9 from the detection source variable of the Florence-2 feature extractor~\citep{xiao2024florence}, a 0.7B vision encoder–decoder model, within the Grounded SAM-2 bounding box generation pipeline. To stabilize the bounding box–grounded MCQ generation process in the Stage 1 pipeline over diverse object-context, we constrain the maximum number of same-category instances per image to 9 (e.g., up to nine ``tree'' bounding box tags in a visual scene image).

\noindent \textbf{Semantic filtering for question diversity.} To ensure the quality and diversity of the generated data, we employ a two-step filtering process. 
First, an Qwen3-30B-A3B-Instruct-2507~\citep{yang2025qwen3} based \textbf{LLM verifier} assesses the factual correctness and logical soundness of each generated MCQ. 
Second, a semantic similarity filtering step is applied to discard redundant or overly similar questions, thereby enhancing the diversity of the final dataset.

For the semantic filtering, we represent each MCQ $i$ by a question stem embedding $q_i$, a selected answer embedding $a_i$, and an optional set of category tags $c_i$. 
Embeddings are derived from all-MiniLM-L6-v2, a 22M text embedding model in SentenceTransformer~\citep{reimers2019sentence}, after standard text normalization. 
We define a composite similarity score between two MCQs, $i$ and $j$, as a weighted combination:
\begin{equation}
   s(i, j) = \lambda_s \cos(q_i, q_j) + \lambda_a \cos(a_i, a_j) + \lambda_c \mathcal{J}(c_i, c_j), 
\end{equation}
where $\cos(\cdot, \cdot)$ denotes cosine similarity and $\mathcal{J}(\cdot, \cdot)$ is the Jaccard index for category tags. 
The weights $\lambda_s$, $\lambda_a$, and $\lambda_c$ balance the contribution of the question stem, answer, and category similarity, respectively.

For each newly generated MCQ $i$, we query an k-nearest neighbors index to retrieve its top-$k$ closest neighbors from the set of already accepted questions. 
We then compute the composite similarity $s(i, j)$ for these neighbors. 
If its maximum similarity to any previously accepted question $j$ 
exceeds the deduplication threshold $\tau_{dup}$:
\begin{equation}
 \max_{j<i} s(i, j) \ge \tau_{dup},
\end{equation}
where $\tau_{dup}$ is a value set to $0.82$ after an ablation study on the filtered question quality. Otherwise, the question is added to our 
filtered dataset and its embedding is added to the index.

\subsection{Additional Analysis of Data Diversity}
\label{suppl:diversity_analysis}

To investigate the source of the scalability gains reported in Section~\ref{suppl:stage_1}, we analyze the semantic diversity of the generated supervision. We hypothesize that conditioning on dense captions alone leads to \textit{MCQ synthesis collapse}, where the generator repeatedly targets the most salient image features (\textit{e.g.}, dominant objects, colors), creating redundant training signals. In contrast, we posit that conditioning on open-vocabulary bounding boxes (LGT) forces the generator to explore the long tail of visual concepts.

We test this by sampling $N=1000$ questions generated from three shared seed images for both LGT (Ours) and LPT (Baseline). We encode these questions using SentenceBERT~\citep{reimers2019sentence} \texttt{all-MiniLM-L6-v2} and estimate their density in the semantic space using Gaussian Kernel Density Estimation (KDE).

{\begin{figure}[h]
    \centering
    \includegraphics[width=0.98\linewidth]{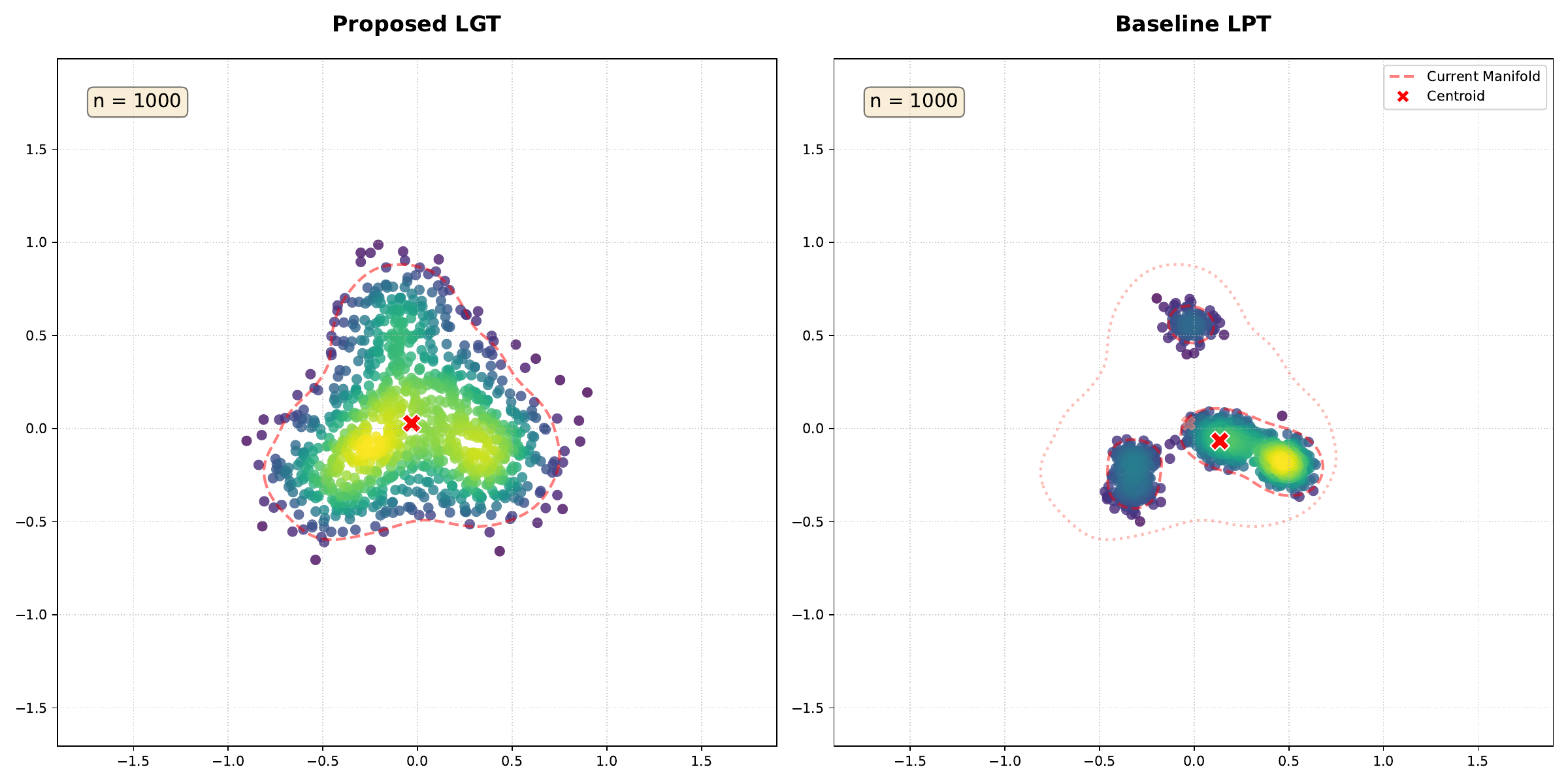}
    \caption{\textbf{Semantic Manifold Comparison.} KDE contours (95\% confidence) reveal that caption-only generation (LPT, right) suffers from mode collapse, clustering tightly around coarse visual features. Object-grounded generation (LGT, left) prevents this collapse, expanding the supervision manifold to cover finer-grained spatial and attribute reasoning. }
    \label{fig:diversity_manifold}
\end{figure}}

As shown in Figure~\ref{fig:diversity_manifold}, the baseline distribution exhibits severe mode collapse, characterized by high density in a narrow semantic region. Quantitatively, LGT demonstrates superior coverage:
\begin{itemize}
    \item \textbf{Semantic Spread:} LGT yields a \textbf{3.2$\times$ wider} distribution (mean Euclidean distance to centroid) than the baseline.
    \item \textbf{Information Density:} LGT significantly reduces redundancy, with an average pairwise cosine similarity of \textbf{0.61} compared to \textbf{0.82} for LPT.
\end{itemize}
These results confirm that object-level metadata prevents MCQ synthesis collapse and maintains diversity at scale. By preventing saturation in the supervision signal, LGT enables the continuous performance scaling observed in our main results, whereas caption-based methods rapidly hit diminishing returns.

\subsection{Vision-Centric Prompt Templates}
\label{ap:s1:prompt}
When the previous LLM based text-to-text MCQs generation frameworks~\citep{liao2025longperceptualthoughts} are highly relied on the image caption quality of the text, injection visual-grounded information has not been well explored. We carefully design a coordinate information guided prompt for the vision-centric MCQ generation with one main task instruction prompt, one core principle, and four auxiliary guidelines as shown in ~\ref{label:prompt:stage:1}. We will open source the full data generation pipeline with the exact prompt instruction.

\begin{promptbox}
\label{label:prompt:stage:1}
You are a computer vision expert generating object-centric visual questions that require precise examination of a specific bounded region. Given a detailed image description and a target object with its bounding box coordinates, create challenging multiple-choice questions that demand careful analysis of the object within its bounded region and its contextual relationships.
\textbf{Generate exactly \texttt{\{\{ num\_questions | default(4) \}\}} multiple-choice questions.}
\end{promptbox}

\noindent \textbf{Prompt Task} At the start of the process, we add a ``computer vision expert'' role-playing instruction; therefore, the text-LLM can better interpret \textit{bounding box coordinates} and image tags while generating questions. In our experiments, using this ``computer vision expert'' instruction led to a noticeable improvement: the instruction-following F1 score rose from $96.3$ to $99.2$ during Stage 1 data generation. Similar role-playing~\citep{zhang2025automated}, self-reflection~\citep{shinn2023reflexion}, and task-activating prompting~\citep{yang2023generative} approaches have also been explored in recent work on domain-specific MCQ generation, such as in medical and speech understanding.

\begin{principlesbox}{Core Principles}
\begin{itemize}
    \item \textbf{Object-Centric Focus}: Every question must center on the specific object within the provided bounding box.
    \item \textbf{Spatial Precision}: Questions should require locating and examining the exact bounded region.
    \item \textbf{Contextual Relationships}: Explore how the target object relates to its surroundings and other elements.
    \item \textbf{Multi-Level Analysis}: Progress from basic object properties to complex spatial and functional relationships.
\end{itemize}
\end{principlesbox}

\noindent \textbf{Core Principle.} To strengthen the model’s ability to reason over complex spatial and visual information in MCQs, we perform an error analysis on MTVQA~\citep{tang2024mtvqa} and MegaBench~\citep{chen2024mega} benchmarks that do not overlap with our test set, using the \texttt{QwenVL-2.5} report as reference. The analysis reveals that most failure cases stem from two sources: 55.1\% misunderstanding spatial relationships (e.g., identifying where an object mentioned in the question is located in the image) and 32.2\% misinterpreting contextual information (e.g., inferring the function or color of the subject). Motivated by these observations, we establish four vision-centric principles to guide question construction, leveraging dense captions, bounding box labels, and coordinate information.

\noindent \textbf{Additional Auxiliary Information.}
We introduce auxiliary rules to further refine question generation in alignment with our core principles. These include guidelines on how to incorporate meta-information when framing vision-centric question categories, strategies to encourage deeper investigation of visual commonsense beyond what dense-caption-based generation typically captures, refinements to both input and output formats, and a final reminder to avoid directly disclosing bounding box coordinates in the question text. Instead, the text should be used to construct the subject that grounds the visual target.

In summary, LGT employs a carefully designed prompt instruction comprising $1,987$ tokens, which is substantially longer than the simpler vision-instruction prompt of $419$ tokens used in ~\cite{liao2025longperceptualthoughts}. The full prompt is provided in detail below.

\begin{instructionbox}{Question Categories}
Distribute your \texttt{\{\{ num\_questions | default(4) \}\}} questions among the following categories:
\begin{enumerate}
    \item \textbf{Specific Region Analysis} (\texttt{\{\{ ((num\_questions | default(4)) * 0.25) | round | int \}\}} question\texttt{\{\{ 's' if ((num\_questions | default(4)) * 0.25) | round | int != 1 else '' \}\}}):
    \begin{itemize}
        \item Object attributes within the bounding box (color, texture, material, size, orientation).
        \item Object state and condition (pose, activity, physical state).
        \item Object parts and components visible within the bounded region.
        \item Visual details that distinguish this object from similar objects.
    \end{itemize}
    \item \textbf{Object-Environment Interactions} (\texttt{\{\{ ((num\_questions | default(4)) * 0.25) | round | int \}\}} question\texttt{\{\{ 's' if ((num\_questions | default(4)) * 0.25) | round | int != 1 else '' \}\}}):
    \begin{itemize}
        \item Spatial relationships between the target object and immediate surroundings.
        \item How the object interacts with or relates to nearby elements.
        \item Environmental context affecting the object's appearance or function.
        \item Lighting, shadows, or reflections involving the target object.
    \end{itemize}
    \item \textbf{Comparative \& Relational Questions} (\texttt{\{\{ ((num\_questions | default(4)) * 0.25) | round | int \}\}} question\texttt{\{\{ 's' if ((num\_questions | default(4)) * 0.25) | round | int != 1 else '' \}\}}):
    \begin{itemize}
        \item How this object compares to other objects in the scene.
        \item Relative positioning, size, or orientation compared to other elements.
        \item Object hierarchies or groupings involving the target object.
        \item Contextual significance of the object within the overall scene.
    \end{itemize}
    \item \textbf{Functional \& Semantic Analysis} (\texttt{\{\{ ((num\_questions | default(4)) * 0.25) | round | int \}\}} question\texttt{\{\{ 's' if ((num\_questions | default(4)) * 0.25) | round | int != 1 else '' \}\}}):
    \begin{itemize}
        \item Object's purpose or function within the scene context.
        \item How the object is being used or what role it plays.
        \item Semantic relationships between the object and scene narrative.
        \item Implied actions or processes involving the target object.
    \end{itemize}
\end{enumerate}
\end{instructionbox}

\begin{principlesbox}{Design Guidelines}
\begin{itemize}
    \item \textbf{Implicit Object Targeting}: Questions should focus on the target object without explicitly revealing bounding box coordinates in the question text.
    \item \textbf{Object Identification Challenge}: Questions must require the reader to first identify and locate the target object based on contextual clues and object description.
    \item \textbf{Progressive Complexity}: Start with direct object attributes, then move to spatial relationships, then to complex contextual analysis.
    \item \textbf{Precise Language}: Use specific spatial terms (e.g., \textit{adjacent to, overlapping with, positioned above}) and descriptive object references.
    \item \textbf{Distractors Strategy}: Create plausible wrong answers that might apply to other objects in the scene but not the target object.
    \item \textbf{Coordinate Disclosure}: \textbf{DO NOT} mention bounding box coordinates in the question text.
    \item \textbf{Design for Multiple-Choice}: Provide 4 answer options (A, B, C, D) with one correct answer and three plausible distractors that require careful inspection to rule out.
    \item \textbf{Clarity, Specificity, and Brevity}: Formulate clear, focused questions that are detailed enough to challenge the reader, avoiding ambiguity or reliance on general knowledge.
\end{itemize}
\end{principlesbox}

\begin{instructionbox}{Input \& Output Format}
\textbf{Image Description:} \texttt{\{\{ image\_description \}\}}

\textbf{Target Object Analysis:}
\begin{itemize}
    \item \textbf{Object}: \texttt{\{\{ bbox\_label \}\}}
    \item \textbf{Image Dimensions}: \texttt{\{\{ image\_width \}\}} $\times$ \texttt{\{\{ image\_height \}\}} pixels
\end{itemize}

\textbf{Structured Output Example:}
\begin{verbatim}
1. <question> Your question here </question>
   <choices> (A)... (B)... (C)... (D)... </choices>
   <answer> object_label, [x1, y1, x2, y2], actual_answer 
   </answer>
   <type> reasoning type here </type>
\end{verbatim}
\end{instructionbox}

\begin{criticalbox}{\faExclamationTriangle~Critical Reminders}
\begin{itemize}
    \item[\faCheckCircle] You must generate \textbf{exactly \texttt{\{\{ num\_questions | default(4) \}\}}} questions.
    \item[\faTimesCircle] Questions \textbf{MUST NOT} disclose bounding box coordinates or specific object labels. Use generic terms like "the object," "the item," or "the element" instead of "\texttt{\{\{ bbox\_label \}\}}".
    \item[\faCheckCircle] Answers \textbf{MUST INCLUDE} the exact object label and integer coordinates in the specified format: \texttt{"\{\{ bbox\_label \}\}, [\{\{ bbox\_coordinates[0]|round|int \}\}, \{\{ bbox\_coordinates[1]|round|int \}\}, ...], [specific answer]"}.
    \item[\faBullseye] Questions must focus on visual properties or relationships of the object within the specified bounded region, requiring careful inspection.
\end{itemize}
\end{criticalbox}

\subsection{Stage 2: Compositional Problem Generation}
\label{appendix:a:3}
While Stage 1 generates a large and diverse set of grounded questions, a significant portion of them are relatively simple and can be solved directly by the base VLM. To push the model's reasoning capabilities further, Stage 2 introduces a \textbf{composition hardening} algorithm designed to systematically increase problem complexity.

The approach is straightforward yet effective: for a given image, we randomly sample up to five question-answer pairs generated in Stage 1. These simpler problems, along with the global image caption, are provided as input to a generator LLM. The LLM is then tasked with composing these individual questions into a single, more challenging multi-hop problem that requires higher-order reasoning to solve. Following generation, we apply a similar filtering protocol as in Stage 1: the generator model is prompted to solve its own composed question, and we retain only those problems with a high answer consistency threshold ($\tau \ge 0.8$), ensuring both difficulty and verifiability. The specific prompt used for this stage is detailed in Appendix \ref{appendix:a:3}.

\begin{promptbox}
You will be given a description of an image and up to 5 different easy problems asking about the image. Use the questions to create a single, creative and hard question to solve.
\begin{itemize}
    \item The question should be in a multiple-choice format with 4 options, just like the given questions.
    \item The composed question should be much harder than each of the individual subquestions provided.
    \item You should focus on the perceptual capabilities (e.g. counting objects, detecting color or texture of an object, relative location of the objects, the angle of the image, detecting letters, etc) and creatively use them to make a harder question.
    \item Do not simply ask about an enumeration of these features, and you may focus on one specific aspect among them. But make sure that the new question is harder to solve.
    \item Only use english in your problem.
\end{itemize}
Your output should be exactly formatted as: \\
\texttt{
Hard problem \\
your hard question \\
(A) your hard problem option A \\
(B) your hard problem option B \\
(C) your hard problem option C \\
(D) your hard problem option D \\
Correct answer: your correct answer \\
}
\end{promptbox}

\noindent \textbf{Simple CoT VLM distillation and Thought Expansion}
A core challenge in synthesizing reasoning traces is that open-weight VLMs often lack the rich, non-linear cognitive behaviors (e.g., subgoal setting, backtracking) seen in frontier LLMs, while human annotation is prohibitively expensive at scale. To address this, we adopt the two-step distillation process from LPT, as illustrated in Figure 1 (bottom). The prompt templates used for this distillation process are adapted from a baseline in~\cite{liao2025longperceptualthoughts}.

First, to ensure the reasoning remains in-distribution for the target model, we prompt a VLM ($\mathcal{M}_{VLM}$) with the image and MCQ to produce a concise initial rationale, termed a ``Simple CoT''. Naively sampling from a powerful reasoning LLM directly often yields out-of-distribution traces that degrade downstream performance. Second, we perform a \textbf{thought expansion} step, where the Simple CoT is used to prime a stronger reasoning LLM ($\mathcal{M}_{Reason}$), which expands upon the visually-grounded trace while injecting more complex problem-solving strategies.

Crucially, our approach scales the model choice with task complexity. For the \textasciitilde{}750K simpler problems in Stage 1, we use efficient models (\texttt{Qwen2.5-VL-7B} and a 32B-scale $\mathcal{M}_{Reason}$). For the more complex compositional problems in Stage 2, we leverage frontier models like \texttt{Qwen2.5-VL-72B} and R1 to generate the richest possible reasoning chains.

\subsection{Local Qwen CoT Verification Prompt}
\label{ap:s:4:prompt}
A critical component of our data synthesis pipeline is ensuring the logical fidelity of the expanded reasoning traces. While the ``thought expansion'' step (Section 2.3) enriches Simple CoTs with complex cognitive behaviors, the powerful reasoning LLM can sometimes produce plausible-sounding but factually incorrect reasoning paths that deviate from the ground-truth answer. To filter these inconsistencies at scale without relying on expensive external judges, we designed the following verification prompt. This prompt tasks a smaller, local model (\texttt{Qwen-32B-A3B-2507-Instruct}) to act as an efficient verifier. The model must assess whether the generated reasoning trace (\texttt{Reflection}) logically supports the known correct \texttt{Answer}, effectively serving as a high-throughput quality control mechanism for our SFT and preference data.

\begin{promptbox}
You will be given a visual question, its answer, and a reflection on an initial thought process. The provided answer is always correct, but the reflection may sometimes be inconsistent with this answer. Your task is to check if the reflection logically and factually supports the provided answer.
\end{promptbox}

\begin{instructionbox}{Verification Process}
You will check for consistency by following these steps:
\begin{enumerate}
    \item \textbf{Understand the Question and the Answer:} First, fully comprehend what is being asked and what the correct final answer is.
    \item \textbf{Derive the Answer Solely from the Reflection:} Carefully read the \texttt{Reflection} text and determine what conclusion it leads to, ignoring the provided \texttt{Answer}.
    \item \textbf{Check for Consistency:} Compare the answer derived from the \texttt{Reflection} (Step 2) with the provided ground-truth \texttt{Answer}.
\end{enumerate}

\tcbline

\textbf{Output Requirement:} At the end of your reasoning, you must answer with \textbf{\boxed{Yes}} if the Reflection is consistent with the Answer; otherwise, answer \textbf{\boxed{No}}.
\end{instructionbox}

\begin{databox}{Input Format}
\noindent\rule{\linewidth}{0.4pt}
\vspace{1mm}

\noindent\textbf{\# Question:} \texttt{<question\_text>} \\
\textbf{Answer:} \texttt{<answer\_text>}

\vspace{2mm}

\noindent\textbf{\# Reflection on the initial thought} \\
\textbf{Reflection:} \texttt{... <last\_30\_words\_of\_reflection>}

\vspace{1mm}
\noindent\rule{\linewidth}{0.4pt}

\end{databox}

\subsection{Additional Experimental Results}
\subsubsection{Multimodal Audio Understanding and Omni Model }
We emphasize our data is completely vision-centric. Here, we conduct experiments on out-of-domain (OOD) complex audio reasoning tasks using the Qwen2.5-Omni-7B-Instruct~\citep{xu2025qwen2} model on the Multimodal Audio Understanding (MMAU) benchmark~\citep{sakshi2024mmau}. Our methodology leverages the modular architecture of Qwen-Omni, which separates the reasoning component (``thinker'') from the answer generation component (``talker''). We first fine-tune only the ``thinker'' dense module using our synthesized data. We keep our modality-specific audio encoder frozen when tuning the backbone dense model. 
After this stage, the fine-tuned ``thinker'' is merged back with the original, pre-trained ``talker'' module.  This strategy aims to enhance the model's intrinsic reasoning capabilities without directly altering the audio-specific knowledge contained within the ``talker'' module, thereby mitigating catastrophic forgetting.

\begin{table}[h!]

\centering

\caption{Omni's generalization Results benefited from Long Grounded Thoughts (LGT) on audio understanding. Best scores in each column are in bold.}

\label{tab:lgt_audio_results}

\begin{tabular}{@{}lccc|c}

\toprule

\textbf{Approach} & \textbf{MMAU-Sound} & \textbf{MMAU-Music} & \textbf{MMAU-Speech} & \textbf{MMAU-avg} \\

\midrule

Gemini-2-Flash & 68.93 & 59.37 & 72.87 & 67.03 \\

GPT-4o & 63.24 & 49.93 & 69.33 & 60.80 \\

\midrule

Qwen2.5-Omni-7B-Instruct & 76.77 & 67.33 & \textbf{68.90} & 71.00 \\

+ Virgo & 64.20 & 59.30 & 64.30 & 62.60 \\

+  LPT & 76.63 & 67.56 & 66.93 & 70.23 \\

\midrule
Ours & & & & \\
+ 130k  & 76.60 & 67.80 & 67.40 & 70.23 \\

+ 250k  & 76.90 & 68.90 & 67.80 & 70.23 \\

+ 500k  & 77.20 & 69.50 & 68.82 & 71.80 \\

+ 1M  & \textbf{78.30} & \textbf{70.20} & 68.80 & \textbf{72.32} \\

\bottomrule

\end{tabular}

\end{table}

The results, presented in Table \ref{tab:lgt_audio_results}, demonstrate a surprising and significant positive transfer. The baseline Qwen2.5-Omni-7B-Instruct already establishes a strong performance with an MMAU average of 71.00, surpassing proprietary models like Gemini-2-Flash (67.03). However, not all reasoning data transfers effectively; fine-tuning on Virgo,a visual math dataset, leads to a substantial performance degradation to 62.60, indicating negative transfer. In contrast, our LGT data shows positive scaling behaviors.

Training on 500k LGTs already pushes the model's performance to 71.80, outperforming the other leading omni-models. By scaling to 1M LGTs, our model achieves a new state-of-the-art average score of 72.32 among the tested configurations. These gains are driven by significant improvements in non-speech acoustic reasoning, with the \textbf{MMAU-Sound} score rising from 76.77 to 78.30 and the \textbf{MMAU-Music} score increasing from 67.33 to 70.20. This result underscores that enhancing core reasoning abilities with our vision-centric data can positively transfer to and improve performance on OOD audio tasks. This complex tracing of learning benefits also extends to more challenging omni-modal tasks involving dual-modal inputs (e.g., vision and audio). Our experimental results with long reasoning traces further support the recent finding~\citep{rouditchenko2025omni} that it is possible to enhance audio content reasoning ability solely by injecting high quality reasoning text data. We provide a qualitative example in \ref{fig:example_audio} on how reasoning traces on recognizing complex and compositional sound and speech events.

\begin{table}[h]
  \centering
  \caption{MMLU-Pro Results}
  \begin{tabular}{lc}
    \toprule
    \textbf{Model} & \textbf{Acc} \\
    \midrule
    Qwen2.5-VL-7B-Instruct & 47.15 \\
    \midrule
    VLAA-thinking & 21.56 \\
    Virgo & 37.95 \\
    \midrule
    LPT  & 50.77 \\
    Ours (SFT Only) & \textbf{50.82} \\
    Ours (SFT + DPO) & 47.07 \\
    Ours (SFT + GRPO) & 47.00 \\
    \bottomrule
  \end{tabular}
  \label{tbl:sup_mmlu_2}
\end{table}

\subsubsection{Text-Only MMLU-PRO }
MMLU-Pro~\citep{wang2024mmlupro} extends the original MMLU~\citep{hendrycks2021measuring} by incorporating more reasoning-intensive questions and enlarging the answer choice set. It covers 14 broad domains (i.e., mathematics, physics, chemistry, and others) and consists of over 12,000 questions in total. In table~\ref{tbl:sup_mmlu_2} we show the extended table that was presented shorted in the main work.

\subsubsection{Embodied Open-Ended QA Benchmark}
\label{app:embodied_section}
We evaluate on a modified version of \emph{Needle in the Embodied Haystack} (NiEH)~\citep{kim2025needlesembodiedhaystackenvironment}, focusing on the \textbf{single-evidence} setting where one frame in a time-ordered trajectory suffices to answer the question. The test set contains  {829} image–question pairs.  In our evaluation,  we restrict the input to a  {2048-token} window  {centered on the ground-truth (answer-bearing) frame}, preserving temporal order so that multiple adjacent frames remain visible around the evidence. 

\textbf{Prompting.} For the Qwen-7B-VL baseline, we use the original paper’s prompt. For our models, we apply a fixed system prompt and the following task instruction: \emph{“These images are the agent's view in time order. Answer the question given the images. Do not include explanation or reasoning in the answer tags. Answer with a single word or words.”}

Table~\ref{tbl:suppl_embqa} summarizes results on the modified NiEH single-evidence benchmark (EM). 

\begin{table}[h]
\centering
\caption{Results on the modified NiEH single-evidence benchmark. Higher is better. Exact Match.}
\label{tbl:suppl_embqa}
\begin{tabular}{l r}
\hline
\textbf{Model} & \textbf{Score} \\
\hline
Qwen2.5-VL-7B-Instruct & 47.55 \\
\midrule
+ VLAA-thinking & 46.75 \\
+ LPT & 51.95 \\
\midrule
+ Ours (SFT Only) & 48.24 \\
+ Ours (SFT + GRPO) & 39.10 \\
+ Ours (SFT + DPO) & \textbf{56.34} \\

\hline
\end{tabular}
\end{table}

\subsection{Experiments on MathVision and VisuLogic Reasoning}
\new{In Figure~\ref{tab:supp_math_visulogic}, we additionally evaluated our best model finetuned on our data on \textbf{MathVision and VisuLogic}. Remarkably, despite our data containing \textbf{no math questions}, we see meaningful improvements on both benchmarks. This validates our data's generalization capability and demonstrates that strengthening core reasoning abilities transfers to out-of-domain tasks.}

\new{
\begin{table}[h]
    \centering
    \caption{Our best model finetuned on our data on \textbf{MathVision and VisuLogic}. Despite our data containing \textbf{no math questions}, we see meaningful improvements on both benchmarks.}
    \label{tab:supp_math_visulogic}
    \vspace{2mm}
    \begin{tabular}{lcc}
        \toprule
        \textbf{Model} & \textbf{MathVision} & \textbf{VisuLogic} \\
        \midrule
        Qwen2.5-VL-7B-Instruct & 49.57 & 11.5 \\
        Ours (SFT and RL) & \textbf{51.77} \small{(+2.2)} & \textbf{19.3} \small{(+7.8)} \\
        \bottomrule
    \end{tabular}
\end{table}
}

\begin{table}[h!]
\centering
\small
 \caption{SFT and RL ablations across data generation algorithms}
\setlength{\tabcolsep}{12pt}
\renewcommand{\arraystretch}{1.1}
\begin{tabularx}{\linewidth}{@{} c c c c c r @{}}
\toprule
\textbf{Data Generation Algo.} & \textbf{Starting Point} & \textbf{SFT Data} & \textbf{RL Algo.} & \textbf{RL Data} & \textbf{Avg Perf.} \\
\midrule
LPT & \basemodel  & 750K & None & None & 0.682 \\
Ours & \basemodel & 750K & None & None & \textbf{0.716} \\
\midrule
\midrule
\multirow{5}{*}{LPT} & \multirow{5}{*}{\basemodel} & \multirow{5}{*}{0} & 
\multirow{5}{*}{GRPO} & 1K & 0.673 \\
& & & & 5K & 0.682 \\
& & & & 10K & \textbf{0.695} \\
& & & & 50K & 0.683 \\
& & & & 100K & 0.669 \\
\midrule
\midrule
\multirow{5}{*}{LPT} & \multirow{5}{*}{SFT} & \multirow{5}{*}{750K} & 
\multirow{5}{*}{GRPO} & 1K & 0.678 \\   
& & & & 5K & 0.658 \\
& & & & 10K & 0.662 \\
& & & & 50K & 0.685 \\
& & & & 100K & \textbf{0.709} \\

\midrule
\multirow{5}{*}{Ours} & \basemodel & 0 & GRPO & 70K & 0.704 \\
 & SFT & 750K & DPO & 70K & 0.737 \\

& SFT & 750K & DPO & 129K & 0.740 \\
& SFT & 750K & GRPO & 70K & \textbf{0.757} \\
& SFT & 750K & GRPO & 129K & 0.746 \\
\bottomrule
\end{tabularx}

\label{tab:rl_sft_ablation_2}
\end{table}

\label{sec:appendix-stats}

\begin{table*}[t]
\centering
\renewcommand{\arraystretch}{1.1}
\caption{\textbf{SFT and RL ablations across data–generation algorithms (Avg over four vision‐centric benchmarks). } Observations:
\textbf{(i)}
Starting from a base instruct model, GRPO peaks at \textbf{0.695} (10K) and degrades at 50–100K, underperforming our SFT baseline \textbf{0.716}; even with an LPT SFT start, GRPO 100K reaches \textbf{0.709} (< \textbf{0.716}). 
\textbf{(ii)}
Staged offline preference learning (\textbf{SFT 750K + DPO 129K} = \textbf{0.740}) is within \textbf{1.7} points of the best online setting (\textbf{SFT 750K + GRPO 70K} = \textbf{0.757}), offering similar accuracy without synchronized RL compute. 
\textbf{(iii)}
GRPO shows early gains then plateaus: best at $\sim$70K on our SFT base (\textbf{0.757}); scaling to 129K reduces to \textbf{0.746}; from a base model, performance declines beyond 10K. 
\textbf{(iv)}
Grounded–MCQ SFT \textbf{0.716} surpasses LPT SFT \textbf{0.682} (+\textbf{3.4} points). 
Bold marks the best in each block; K = thousands.}

\begin{tabularx}{0.8\linewidth}{@{} c c c c c c @{}}
\toprule
\textbf{Data Generation Algo.} & \textbf{Starting Point} & \textbf{SFT Data} & \textbf{RL Algo.} & \textbf{RL Data} & \textbf{Avg Perf.} \\
\midrule
LPT & \basemodel  & 750K & None & None & 0.682 \\
Ours & \basemodel & 750K & None & None & \textbf{0.716} \\
\midrule
\midrule
\multirow{3}{*}{LPT} & \multirow{3}{*}{\basemodel} & \multirow{3}{*}{0} & 
\multirow{3}{*}{GRPO} & 10K & \textbf{0.695} \\
& & & & 50K & 0.683 \\
& & & & 100K & 0.669 \\
\midrule
\midrule
\multirow{3}{*}{LPT} & \multirow{3}{*}{SFT} & \multirow{3}{*}{750K} & 
\multirow{3}{*}{GRPO} & 10K & 0.662 \\  
& & & & 50K & 0.685 \\
& & & & 100K & \textbf{0.709} \\

\midrule
\multirow{5}{*}{Ours} & \basemodel & 0 & GRPO & 70K & 0.704 \\
 & SFT & 750K & DPO & 70K & 0.737 \\

& SFT & 750K & DPO & 129K & 0.740 \\
& SFT & 750K & GRPO & 70K & \textbf{0.757} \\
& SFT & 750K & GRPO & 129K & 0.746 \\
\bottomrule
\end{tabularx}
%
\label{tab:rl_sft_ablation}
\end{table*}

\begin{table}[h]
\centering
\small
\setlength{\tabcolsep}{5pt}
\caption{\textbf{Impact of CoT Richness (Stage 1 Data).} Simple traces (Short CoT) hurt performance; only rich, expanded traces (Full CoT) yield best results.}
\label{tab:cot_ablation}
\vspace{-2mm}
\begin{tabular}{lcc}
\toprule
\textbf{Configuration} & \textbf{CV-Bench} & \textbf{V*Bench} \\
\midrule
Base (Qwen2.5-VL-7B) & 0.740 & 0.485 \\
+ Thinking Prompt (No SFT) & 0.754 & 0.551 \\
+ SFT (No CoT) & 0.787 & 0.583 \\
+ SFT (Short CoT) & 0.714 & 0.540 \\
\textbf{+ SFT (Full CoT - Ours)} & \textbf{0.813} & \textbf{0.597} \\
\bottomrule
\end{tabular}
\end{table}

\begin{table}[t!]
\centering
\caption{\rebuttal{Detailed cognitive behaviour statistics for various datasets.} Values represent the average count of the behaviour on a subsample of 1000 examples  of the dataset. Quantification methodology and terminology follows from ~\cite{gandhi2025cognitivebehaviorsenableselfimproving,liao2025longperceptualthoughts}.}
\begin{adjustbox}{width=\linewidth}
\begin{tabular}{@{}lccccccccc@{}}
\toprule
\textbf{Behaviour} & 
\textbf{\shortstack{LongPerceptual\\Thoughts}} & 
\textbf{\shortstack{LongGrounded\\Thoughts (s1)}} & 
\textbf{\shortstack{LongGrounded\\Thoughts (s2)}} & 
\textbf{\shortstack{allenai/pixmo\\-ask-model}} & 
\textbf{DriveLMMo1} & 
\textbf{LENS} & 
\textbf{SCI-Reason} & 
\textbf{VLLA-Thinking} & 
\textbf{Virgo} \\
\midrule

Subgoal setting & 0.036 & 0.12 & 0.55 & 0.017 & 0.235 & 0 & 0.032 & 0.17 & 0.24 \\
Backtracking    & 0.354 & 0.35 & 0.68 & 0     & 0     & 0 & 0     & 0.37 & 0.79 \\
Verification    & 0.26  & 0.33 & 0.76 & 0     & 0     & 0 & 0     & 0.45 & 0.89 \\
\midrule
Sum & 0.65&	0.8&	1.99&	0.017&	0.235&	0	&0.032&	0.99&	1.92 \\
\bottomrule
\end{tabular}%
\end{adjustbox}
\label{tab:appendix-stats-detailed-behaviours}
\end{table}

\begin{figure}[t!]
    \centering
    \includegraphics[width=0.97\linewidth]{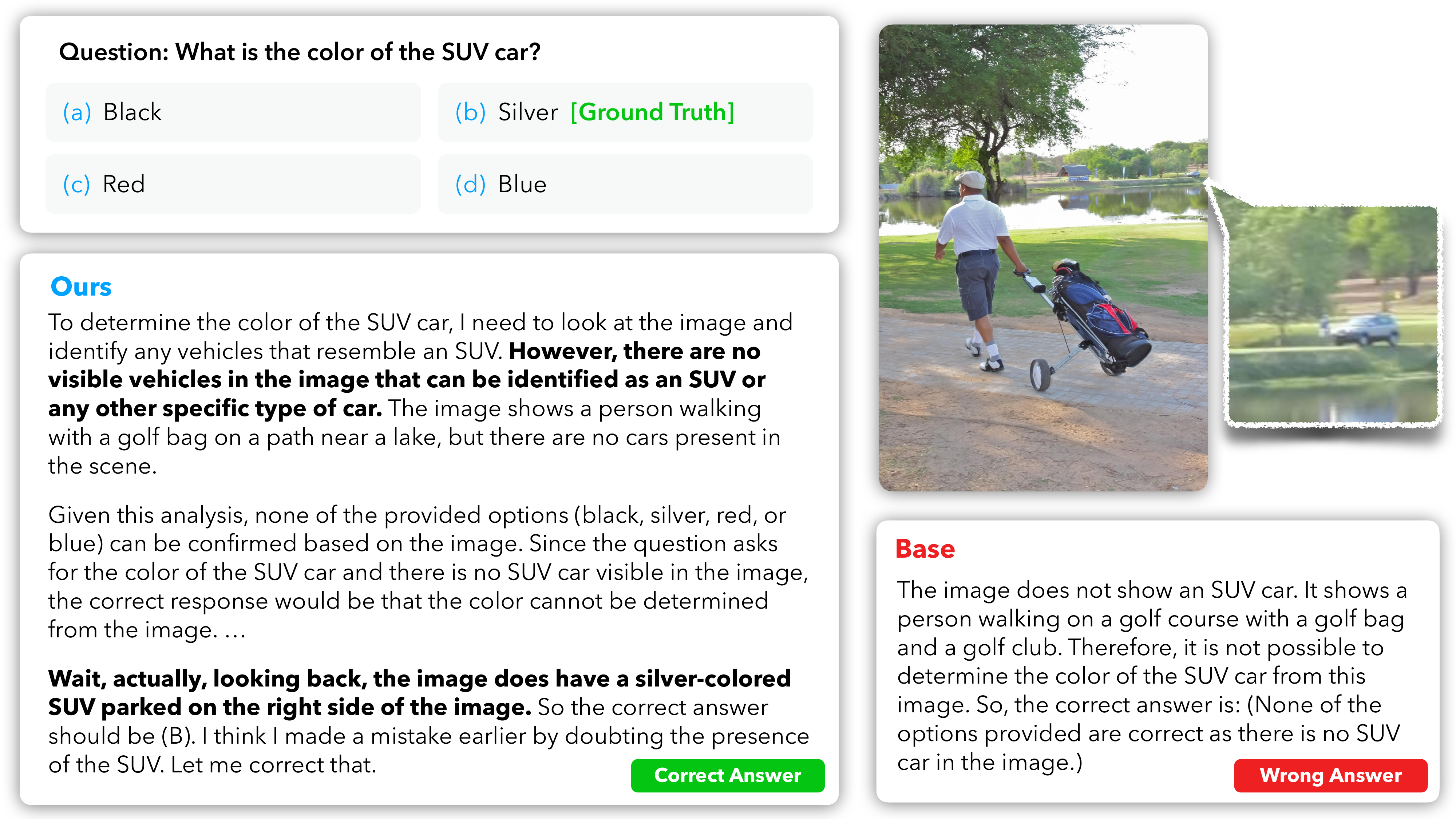}
    \caption{\rebuttal{Additional qualitative example of a reasoning trace from the  model post-trained on our data vs the base model.}}
    \label{fig:example_suv}
\end{figure}

\begin{figure}[t!]
    \centering
    \includegraphics[width=0.97\linewidth]{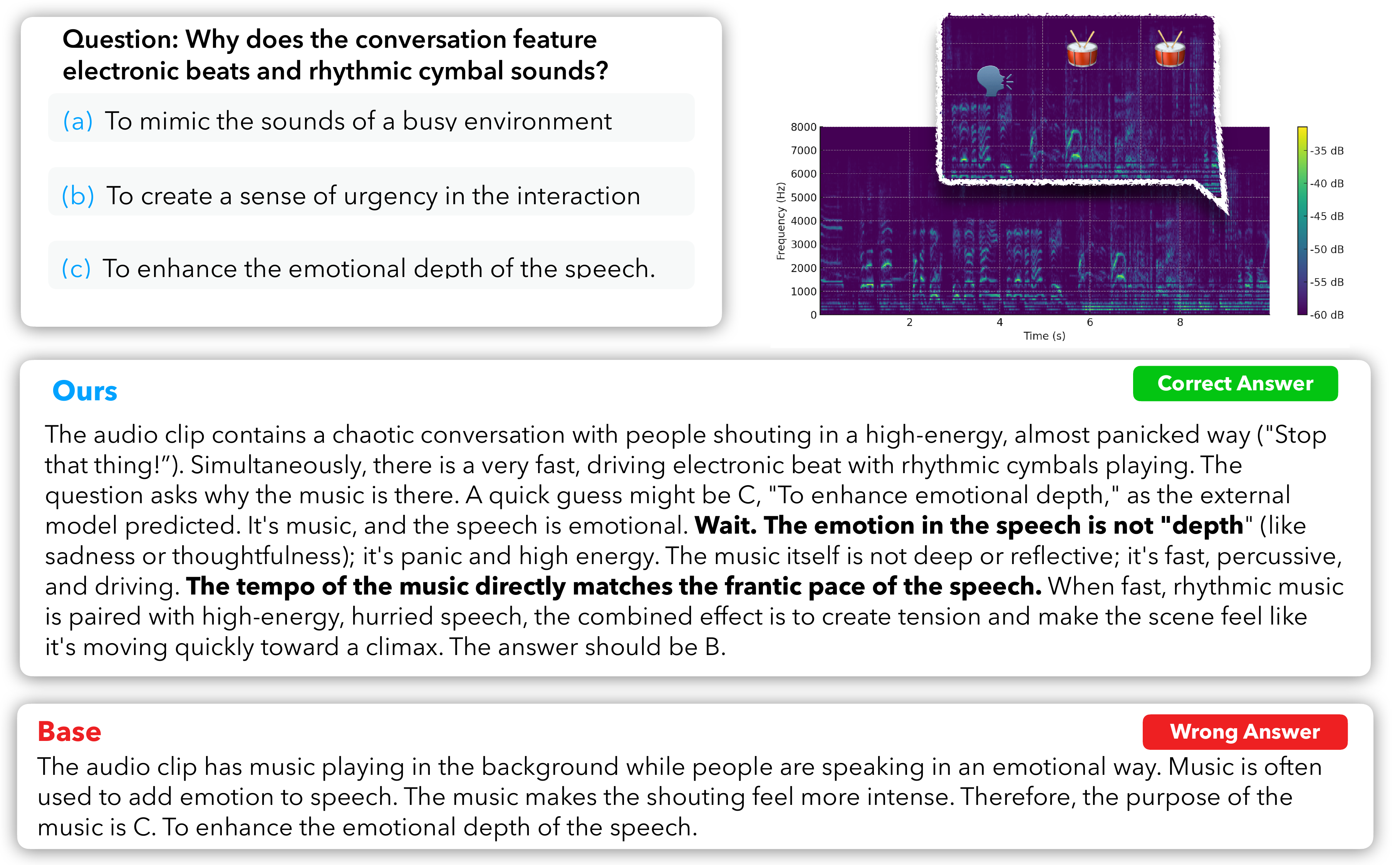}
    \caption{\rebuttal{Temporal reasoning improvement illustrated by a qualitative example of a reasoning trace from the Qwen-2.5 Omni model post-trained on our LGT data, compared to the base Qwen-2.5 Omni model, on an unseen audio reasoning question involving joint speaking and sound events recognition.}}
    \label{fig:example_audio}
\end{figure}